%% file: main.tex
\documentclass[12pt]{article}

\usepackage{newtxtext,newtxmath}

\usepackage{graphicx}
\usepackage{orcidlink}
\usepackage[utf8]{inputenc}
\usepackage[T1]{fontenc}
\usepackage{hyperref}
\usepackage{breakurl}
\usepackage{booktabs}

\usepackage[letterpaper,margin=1in]{geometry}
\usepackage{cprotect}
\usepackage{subfig}
\usepackage{tikz}
\usepackage{graphicx}
\usepackage{multirow}
\usepackage{float}
\usepackage{wrapfig}
\usepackage{xspace}
\usepackage{lineno}
\usetikzlibrary{positioning}

\usepackage{tikz}
\usetikzlibrary{positioning, fit}

\input{macros}

\renewenvironment{abstract}
	{\quotation}
	{\endquotation}

\date{}

\makeatletter
\renewcommand{\fnum@figure}{\textbf{Figure \thefigure}}
\renewcommand{\fnum@table}{\textbf{Table \thetable}}
\makeatother

\usepackage{scicite}

\usepackage{url}

\def\scititle{
	Modular Deep-Learning-Based Early Warning System for Deadly Heatwave Prediction
}
\title{\bfseries \boldmath \scititle}

\author{
	Shangqing Xu$^{1\ast}$,
	Zhiyuan Zhao$^{1}$,
    Megha Sharma$^{1}$,
    José María Martín-Olalla$^{2}$, \and
    Alexander Rodríguez$^{3}$,
    Gregory A. Wellenius$^{4}$,
    B. Aditya Prakash$^{1\ast}$\and
	\small$^{1}$College of Computing, Georgia Institute of Technology, Atlanta, GA, USA.\and
	\small$^{2}$Departamento de Física de la Materia Condensada, Facultad de Física, Universidad de Sevilla, Sevilla, Spain.\and
    \small$^{3}$Computer Science and Engineering, University of Michigan, Ann Arbor, MI, USA.\and
    \small$^{4}$Department of Environmental Health, School of Public Health, Boston University, Boston, MA, USA.\and
	\small$^\ast$Corresponding authors. Email: sxu452@gatech.edu, badityap@cc.gatech.edu\and
}

\begin{document} 

\maketitle


\input{sections/abstract}


\input{sections/intro}


\input{sections/results}

\input{sections/discussion}

\input{sections/method}


\clearpage 

%
\bibliography{main} 
\bibliographystyle{sciencemag}

%
%
%
%
%
%


\paragraph*{Author contributions:}

S.X., Z.Z., J.M.M.-O., A.R., G.A.W., B.A.P. conceived the experiments, S.X., Z.Z., M.S. conducted the experiments, S.X., Z.Z., J.M.M.-O., A.R., G.A.W., B.A.P. analyzed the results, S.X., Z.Z., J.M.M.-O., A.R., G.A.W., B.A.P. reviewed the manuscript.

\paragraph*{Competing interests:}
There are no competing interests to declare.
\paragraph*{Data and materials availability:}
Our codes are available in \url{https://github.com/SigmaTsing/DeepTherm}. The used mortality and non-mortality data should be acquired upon proper request to MoMo, INE database, Aemet Opendata, and Spatial Synoptic Classification V3.0, respectively.


\subsection*{Supplementary materials}
Supplementary Text\\
Tables S1 to S4\\
Figure S1\\


\newpage


\renewcommand{\thefigure}{S\arabic{figure}}
\renewcommand{\thetable}{S\arabic{table}}
\renewcommand{\theequation}{S\arabic{equation}}
\renewcommand{\thepage}{S\arabic{page}}
\setcounter{figure}{0}
\setcounter{table}{0}
\setcounter{equation}{0}
\setcounter{page}{1} 


\begin{center}

\section*{Supplementary Materials for\\ \scititle}

Shangqing Xu$^{1\ast}$,
Zhiyuan Zhao$^{1}$,
Megha Sharma$^{1}$,
José María Martín-Olalla$^{2}$, \\
Alexander Rodríguez$^{3}$,
Gregory A. Wellenius$^{4}$,
B. Aditya Prakash$^{1\ast}$\\ 
\small$^\ast$Corresponding authors. Email: sxu452@gatech.edu, badityap@cc.gatech.edu\\
\end{center}

\subsubsection*{This PDF file includes:}
Supplementary Text\\
Tables S1 to S4\\
Figure S1\\

\newpage

\input{sections/supplementary}



\end{document}

%% file: macros.tex
\iftrue
    \newcommand{\xsq}[1]{\textcolor{purple}{[Shangqing: #1]}}

\else
    \newcommand{\xsq}[1]{}
\fi

\newcommand{\model}{\textsc{DeepTherm}\xspace}

%% file: sections/abstract.tex
\begin{abstract} \bfseries \boldmath
Severe heatwaves in urban areas significantly threaten public health, calling for establishing early warning strategies. Despite predicting occurrence of heatwaves and attributing historical mortality, predicting an incoming deadly heatwave remains a challenge due to the difficulty in defining and estimating heat-related mortality. 
Furthermore, establishing an early warning system imposes additional requirements, including data availability, spatial and temporal robustness, and decision costs.
To address these challenges, we propose \model, a modular early warning system for deadly heatwave prediction without requiring heat-related mortality history. By highlighting the flexibility of deep learning, \model employs a dual-prediction pipeline, disentangling baseline mortality in the absence of heatwaves and other irregular events from all-cause mortality. 
We evaluated \model on real-world data across Spain. Results demonstrate consistent, robust, and accurate performance across diverse regions, time periods, and population groups while allowing trade-off between missed alarms and false alarms.
\end{abstract}

%% file: sections/intro.tex
\section{Introduction}
\noindent
Severe heatwaves, characterized by consecutive days of extremely high temperatures \cite{kysely2010recent, pezza2012severe, murari2015intensification}, are a significant driver of excess mortality in urban areas and pose a major threat to public health \cite{robine2008death, barriopedro2011hot, gasparrini2011impact, ballester2023heat}. For instance, the 2003 European heatwave caused over 40,000 deaths \cite{garcia2010review}, including approximately 2,000 in the Netherlands \cite{garssen2005effect} and 3,000 in Spain \cite{simon2005mortality}. Mitigating the impact of these events necessitates their accurate prediction and the development of effective early warning systems \cite{brimicombe2024preventing}.

Although extensive research has focused on predicting the occurrence of high temperatures \cite{chattopadhyay2011univariate, abhishek2012weather, mckinnon2016long, prodhomme2022seasonal} or attributing historical mortality to heat \cite{gasparrini2011impact, gasparrini2015mortality, kang2018north, vicedo2021burden}, forecasting whether an incoming heatwave will be deadly or not remains a significant challenge \cite{ka2021deadly, ballester2023heat}. A primary obstacle lies in the difficulty of obtaining suitable data. An accurate deadly heatwave prediction model requires data on excess mortality—the number of deaths above a baseline level that are attributable to heat—but this metric is complex to define and estimate precisely \cite{vicedo2021burden, ballester2023heat}. Official records typically only provide all-cause mortality, which combines this desired excess mortality with baseline mortality (deaths from other causes). This fundamental data limitation complicates the development of reliable deadly heatwave predictors.


Among all the existing progress around deadly heatwaves, only a few approaches propose to circumvent the challenge of acquiring direct heat-related mortality data. The proposed methods include projecting non-mortality variables such as meteorology and population demographics to predict future excess mortality\cite{mathes2017real}, or utilizing long-term-averaged mortality as a reference value \cite{mistry2024real}. Yet, they both suffer from key limitations. First, by only modeling the relationship from non-mortality variables or long-term averages to daily mortality in the future, they cannot utilize the rich predictive information contained within daily historical mortality data. Second, these models still require ground-truth excess mortality data for calibration and evaluation, which, as previously established, is difficult to obtain in practice.

Beyond the initial excess mortality prediction, implementing an effective early warning system for deadly heatwaves presents further practical challenges. First, since the system’s outputs are designed to inform policy decisions, the model must be optimized to balance the significant costs associated with both false alarms and missed detections \cite{stainforth2007confidence, ripberger2015false, leclerc2015cry}. Second, the system must be robust and generalizable, maintaining its reliability when deployed in different regions or as new data becomes available. This transportability is crucial for its broader application and impact \cite{burgeno2020impact}.

In summary, developing an early warning system for deadly heatwave classification presents multiple challenges: (1) the inherent difficulty of predicting deadly heatwaves among all incoming heatwaves due to a lack of heat-related mortality data, and (2) the requirements of data availability, robustness, and consideration of decision-related risks. 
Existing early warning systems for heatwaves remain limited to categorizing heatwaves by applying temperature thresholds derived from historical temperature-mortality relationships \cite{sheridan2003heat, park2020heatwave,  hondula2022spatial, li2023regional, martin2024heat}. To the best of our knowledge, no existing studies have developed or validated early warning systems to reveal the presence of deadly heatwaves based on mortality predictions.


To tackle these challenges, we propose a novel dual-prediction strategy to address the challenge of data scarcity. Our approach concurrently predicts two quantities: (1) all-cause mortality and (2) baseline mortality (i.e., mortality in the absence of irregular events like heatwaves). This strategy is motivated by the observation that baseline mortality is far less sensitive to short-term factors, such as daily weather, than all-cause mortality, which has been leveraged by previous mortality attribution studies \cite{weinberger2020estimating, lo2022estimating}. Accordingly, we propose to use a simpler model trained on long-term historical data to capture the stable baseline mortality, while a more complex model uses short-term records and non-mortality factors to predict the all-cause mortality. The difference between these two predictions yields an estimate of excess mortality. Although this value may include effects from other non-heat-related events, it serves as a robust proxy for identifying and classifying deadly heatwaves. Inside such a design, predicting all-cause mortality requires analyzing the complex relationship between mortality and non-mortality variables. Therefore, we propose to introduce deep-learning-based models as the all-cause predictor, leveraging their strong capabilities in complex sequence prediction \cite{wu2022timesnet, kim2019deep}. 

\begin{figure}[ht]
    \centering
    \includegraphics[width=\linewidth]{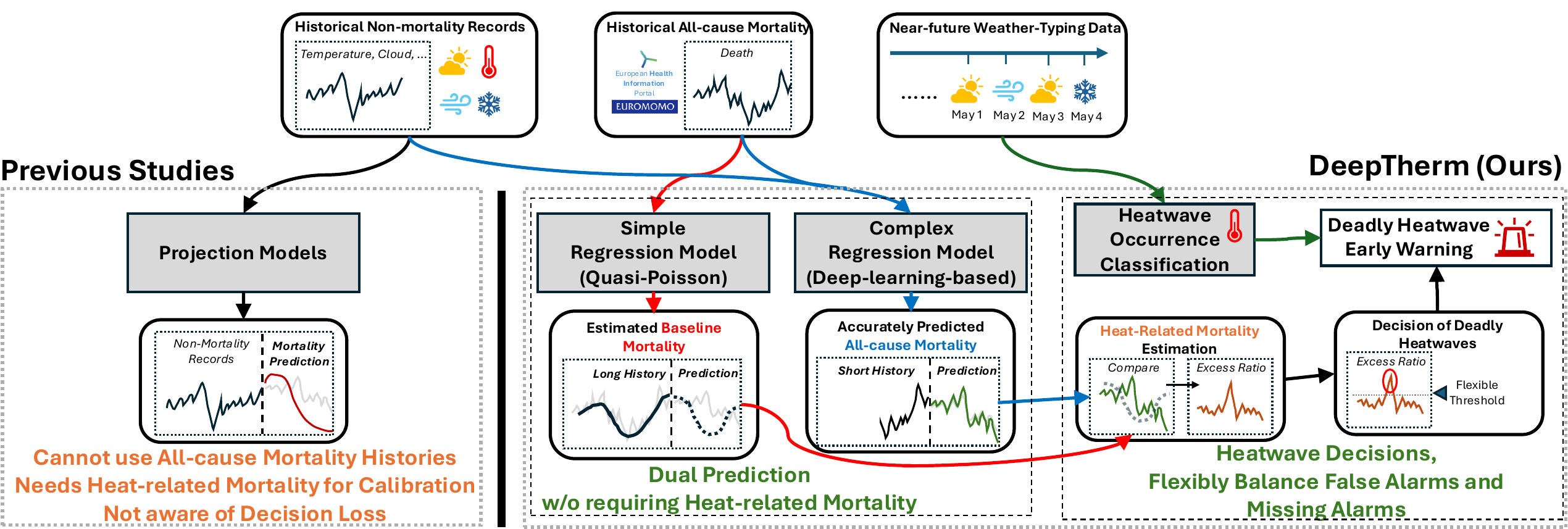}
    \caption{\textbf{\model is capable of using all-cause mortality history with non-mortality records to yield early warnings for deadly heatwaves.} Existing studies of deadly heatwaves remain on projecting non-mortality variables to heat-related mortality, which cannot use historical all-cause mortality records and need heat-related mortality for calibration. Our system, \model, bypasses this challenge with a novelly proposed dual-prediction pipeline. Using historical all-cause mortality data, historical non-mortality data, and near-future synoptic weather-typing data, \model isolates heat-related and baseline mortality components, enabling deadly heatwave identification without requiring heat-related mortality data. 
    Furthermore, \model determines whether an incoming heatwave will be deadly via a flexible thresholding strategy, capable of balancing false alarms and missed alarms.}
    \label{fig:overview}
\end{figure}

Based on our dual-prediction strategy, we establish \model, a modular early warning system for deadly heatwaves, as illustrated in Figure \ref{fig:overview}. \model is composed of three key components: (1) a deep learning module that forecasts all-cause mortality; (2) a Quasi-Poisson regression module that predicts baseline mortality, thereby enabling the estimation of excess mortality; and (3) a decision module that uses this estimate to classify whether a heatwave qualifies as deadly. The deep-learning-based prediction module is expected to accurately predict the all-cause mortality data given historical all-cause mortality data and historical non-mortality data, which is accessible in most regions through national agencies—ensuring broad temporal applicability. Furthermore, we introduce synoptic weather-typing data that categorize weather patterns linked to heatwave occurrences \cite{pezza2012severe, ventura2023analysis}. Due to advances in synoptic prediction systems \cite{sheridan2002redevelopment, hondula2014ssc, cheng2011synoptic, siegert2017synoptic}, these data are widely available for the near-future, and we therefore incorporate their outputs in \model as references for future heatwave occurrences. Additionally, \model employs a flexible thresholding strategy for issuing deadly heatwave alarms, allowing adjustments based on varying tolerance levels for false alarms versus missed detections.

We evaluated \model's performance across 12 provinces of Spain (2015–2023) and their corresponding capital cities (1995–2023). In each city and province, we assess \model in a \textit{real-time} manner: on each day $t$, we train \model with all available non-mortality and mortality data up to day $t$ and synoptic weather-typing data up to day $t+5$ to give deadly heatwave predictions from day $t+1$ to day $t+5$. \model processes daily non-mortality variables (e.g., temperature, humidity, and wind speed) alongside historical all-cause mortality data to classify heatwaves into three categories: level 0, 1, and 2, based on how much the heat-related mortality accounts for in the all-cause mortality. This categorization setting allows us to quantify \model's performance under broader or stricter definitions of deadly heatwaves.
Our results demonstrate that \model achieves: $80\%$ detection rate for level 1 heatwaves with a less than $15\%$ false alarm rate, and $60\%$ detection rate for level 2 heatwaves with a $25\%$ false alarm rate. Moreover, the system exhibits consistent robustness across different age groups (including younger and older populations), evaluation years, and geographic regions with varying climate conditions. Furthermore, we show that \model supports customizable alarm thresholds, enabling policymakers to balance the trade-offs between missed alarms and false alarms according to operational needs.


%% file: sections/results.tex
\section{Results}

\subsection{Experiment Settings}

For a multi-day heatwave, if heat-related mortality exceeds $15\%$ 
of baseline mortality on any day, we classify it as a Level 1 (L1) heatwave; and if heat-related mortality exceeds $30\%$ of baseline mortality on any day, we further classify it as a Level 2 (L2) heatwave; otherwise, the heatwave is categorized as a Level 0 heatwave. Given the time series of historical all-cause mortality and non-mortality data for a given location, \model predicts for each incoming heatwave: 1) whether it will surpass \textit{Level 1}, and 2) whether it will surpass \textit{Level 2}. The occurrence of incoming heatwaves is predicted by synoptic weather-typing data, where we choose Spatial Synoptical Classifications \cite{sheridan2002redevelopment} (Details shown in Section \ref{subsec:problemSetup}). 

To ensure sufficient training data, we begin the evaluation from the third year in each city/province. Additionally, to reduce computational costs, we update mortality predictors annually while retaining the previous year's weights for the all-cause mortality predictor. Meanwhile, to maintain the accuracy of the baseline mortality predictor, we re-train the Quasi-Poisson model from scratch each year using only the two most recent years' data, as population changes over time would otherwise alter the distribution of baseline mortality. By default, we maintain thresholds of \model in $15\%$ for L1 heatwaves and $30\%$ for L2 heatwaves throughout the study. The sources of data can be found in Section \ref{subsec:dataset}. Original result tables will be shown in Supplementary.

\subsection{Spatial Robustness: Performance across Cities and Provinces} \label{subsec:mainResults}

\begin{table}[!t]
    \centering
    \subfloat[Scores across all areas in classifying provincial heatwaves]{\resizebox{\linewidth}{!}{\begin{tabular}{c|c|cccccccccccc|c} \toprule
 & City & Alicante & Badajoz & Barcelona & Biscay & Cordoba & Madrid & Malaga & Orense & Seville & Toledo & Valencia & Zaragoza & \textbf{Total} \\ \midrule
\multirow{4}{*}{L1} & Acc. & 100.0 & 87.8 & 50.0 & 50.0 & 84.9 & 80.0 & 76.9 & 83.3 & 79.0 & 72.7 & 88.9 & 68.4 & \textbf{77.0} \\
 & Pr. & 50.0 & 88.6 & 66.7 & 100.0 & 91.9 & 84.4 & 85.7 & 83.3 & 82.9 & 100.0 & 85.7 & 75.0 & \textbf{80.6} \\
 & Rec. & 100.0 & 97.5 & 66.7 & 100.0 & 87.2 & 84.4 & 75.0 & 100.0 & 90.2 & 70.0 & 100.0 & 85.7 & \textbf{87.4} \\
 & F1 & 100.0 & 92.9 & 66.7 & 66.7 & 89.5 & 84.4 & 80.0 & 50.9 & 86.4 & 82.4 & 92.3 & 80.0 & \textbf{83.8} \\ \midrule
\multirow{4}{*}{L2} & Acc. & 100.0 & 57.1 & 50.0 & 50.0 & 73.6 & 68.0 & 76.9 & 66.7 & 69.4 & 45.5 & 44.4 & 57.9 & \textbf{65.1} \\
 & Pr. & - & 57.6 & 0.0 & - & 73.9 & 42.9 & 25.0 & 75.0 & 66.7 & 37.5 & 40.0 & 63.6 & \textbf{58.3} \\
 & Rec. & - & 73.1 & 0.0 & 0.0 & 68.0 & 42.9 & 100.0 & 75.0 & 84.2 & 75.0 & 50.0 & 63.6 & \textbf{67.3} \\
 & F1 & - & 64.4 & - & - & 70.8 & 42.9 & 40.0 & 75.0 & 74.4 & 50.0 & 44.4 & 63.6 & \textbf{62.4} \\ \bottomrule
\end{tabular}}\label{subfig:momoTable}} \\
    \subfloat[Scores across all cities in classifying city-level heatwaves]{\resizebox{\linewidth}{!}{\begin{tabular}{c|c|cccccccccccc|c} \toprule
 & City & Alicante & Badajoz & Barcelona & Bilbao & Cordoba & Madrid & Malaga & Orense & Seville & Toledo & Valencia & Zaragoza & \textbf{Total} \\ \midrule
\multirow{4}{*}{L1} & Acc. & 78.6 & 76.8 & 77.8 & 66.7 & 82.3 & 68.1 & 61.9 & 70.0 & 73.5 & 76.0 & 88.9 & 80.0 & \textbf{74.9} \\
 & Pr. & 90.0 & 82.5 & 100.0 & 66.7 & 82.6 & 68.0 & 64.7 & 70.0 & 78.7 & 86.4 & 92.9 & 90.2 & \textbf{78.7} \\
 & Rec. & 81.8 & 91.2 & 71.4 & 100.0 & 97.9 & 86.7 & 84.6 & 100.0 & 87.6 & 86.4 & 92.9 & 84.1 & \textbf{89.6} \\
 & F1 & 85.7 & 86.7 & 83.3 & 80.0 & 89.6 & 76.2 & 73.3 & 82.4 & 82.9 & 86.4 & 92.9 & 87.1 & \textbf{83.8} \\ \midrule
\multirow{4}{*}{L2} & Acc. & 78.6 & 72.5 & 66.7 & 83.3 & 66.9 & 75.9 & 66.7 & 40.0 & 63.6 & 68.0 & 77.8 & 63.6 & \textbf{69.2} \\
 & Pr. & 80.0 & 81.5 & 66.7 & 100.0 & 69.6 & 47.8 & 50.0 & 37.5 & 61.7 & 66.7 & 80.0 & 73.1 & \textbf{66.6} \\
 & Rec. & 88.9 & 83.0 & 50.0 & 75.0 & 83.1 & 64.7 & 71.4 & 75.0 & 59.7 & 93.3 & 57.1 & 59.4 & \textbf{73.1} \\
 & F1 & 84.2 & 82.2 & 57.1 & 85.7 & 75.7 & 55.0 & 58.8 & 50.0 & 60.7 & 77.8 & 66.7 & 65.5 & \textbf{69.7} \\
 \bottomrule
\end{tabular}}\label{subfig:ineTable}}
    \caption{\textbf{\model consistently performs in cities and provinces located across Spain, both in classifying L1 heatwaves and L2 heatwaves.} Tables summarize \model's prediction results for (a) provincial and (b) city-level heatwaves, including: 1) Accuracy (Acc.), 2) Precision (Pr.), 3) Recall (Rec.), and 4) F1 score (definitions are provided in the Supplementary Materials). All values are reported as percentages. On average, \model achieves approximately $75\%$ Precision and $85\%$ Recall for L1 heatwaves, and $55\%$ Precision and $65\%$ Recall for L2 heatwaves. It misses fewer than 5 out of 20 L1 heatwaves and 7 out of 20 L2 heatwaves, while generating only 3 false alarms for L1 and 7 for L2 out of 20 alarms. The model’s robust performance across cities and provinces under diverse climate conditions confirms its ability to effectively capture the relationship between non-mortality data and all-cause mortality. }
    \label{fig:spatialResult}
\end{table}

\begin{figure}[!t]
    \centering
    \subfloat[Map chart of provincial prediction accuracies on L1 heatwaves]{\includegraphics[width = 0.47\linewidth]{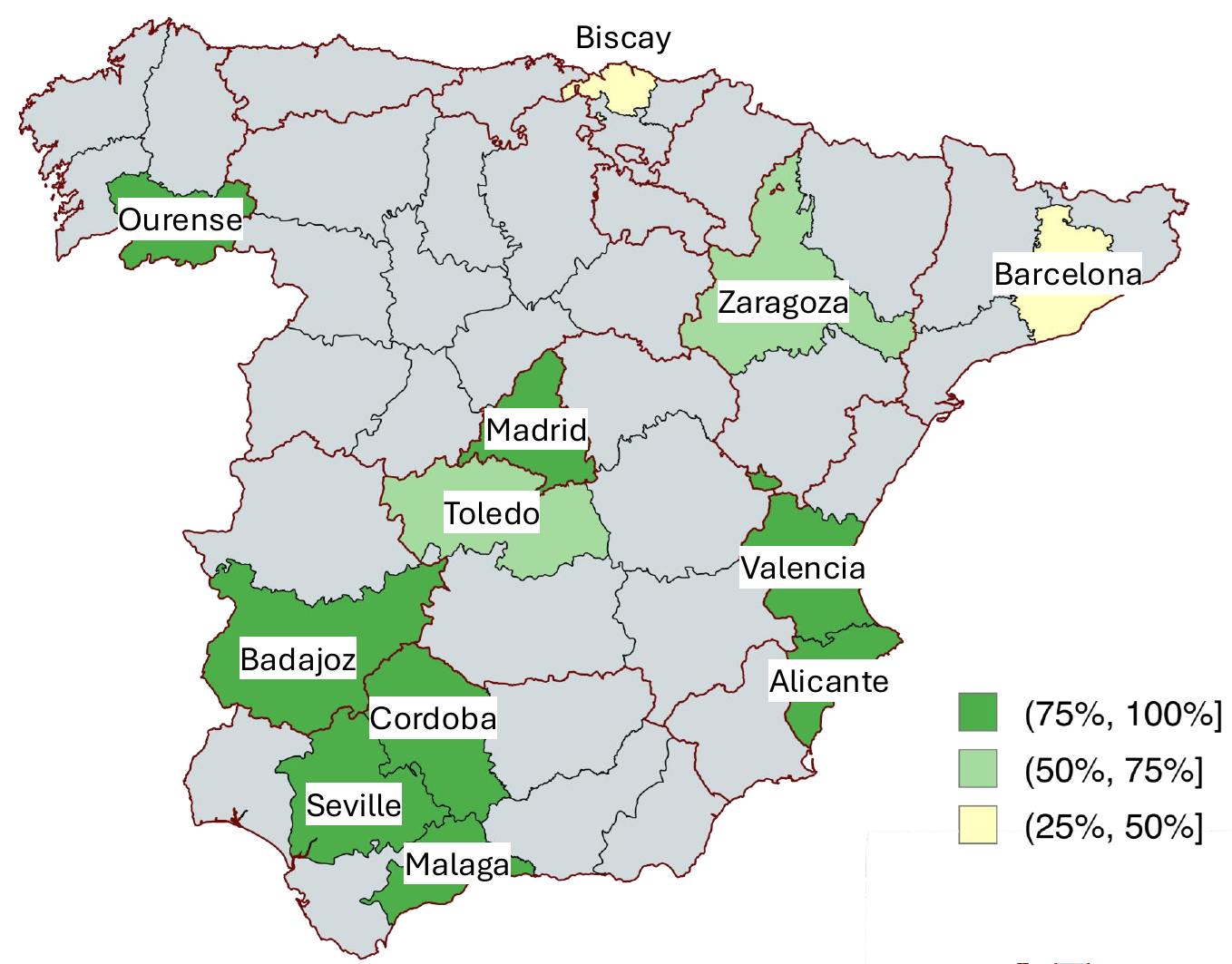}\label{subfig:normalHeatmap}} \hfill
    \subfloat[Map chart of provincial prediction accuracies on L2 heatwaves
    ]{\includegraphics[width = 0.47\linewidth]{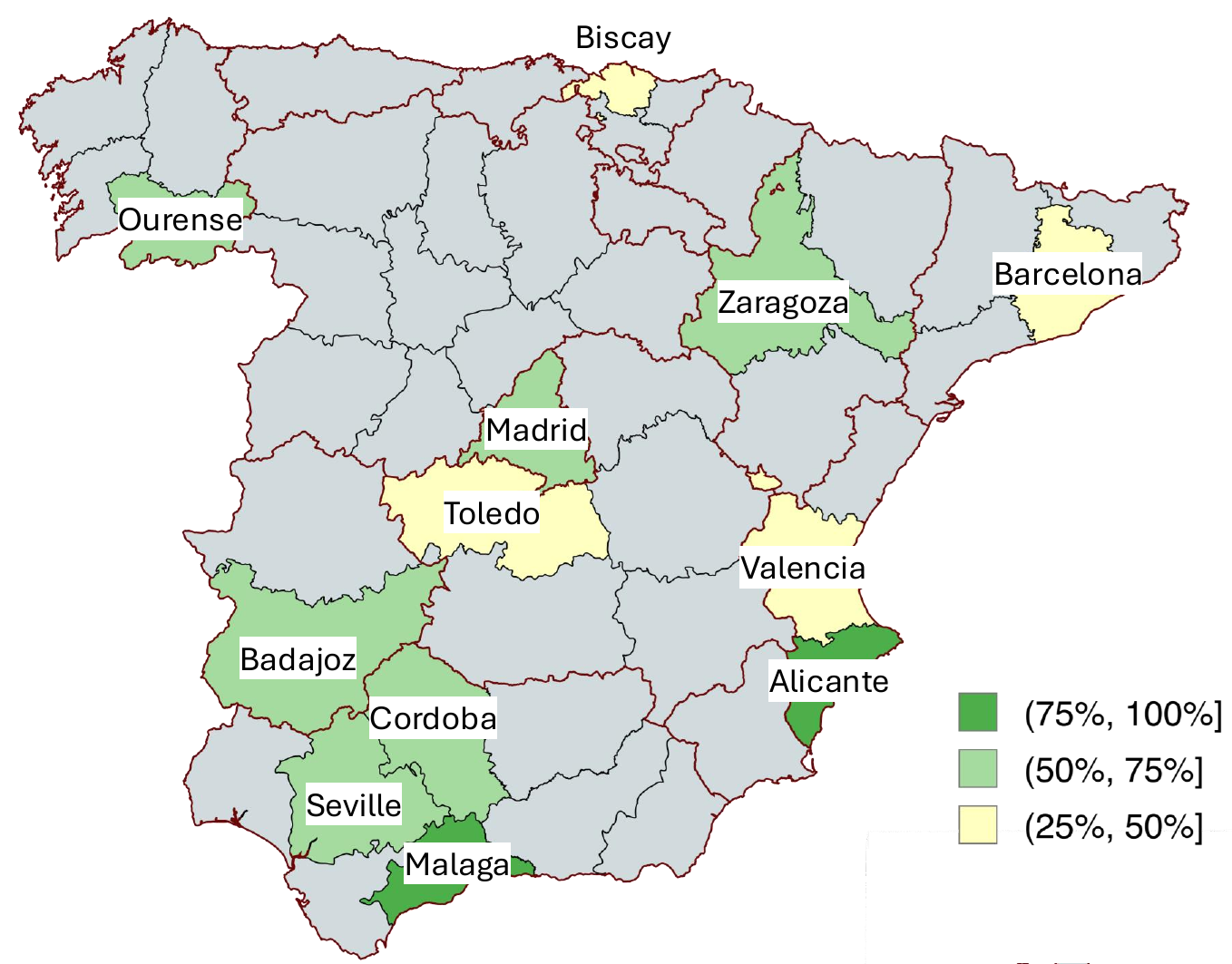}\label{subfig:extremeHeatmap}} \\
    \subfloat[Map chart of city-level prediction accuracies on L1 heatwaves]{\includegraphics[width = 0.47\linewidth]{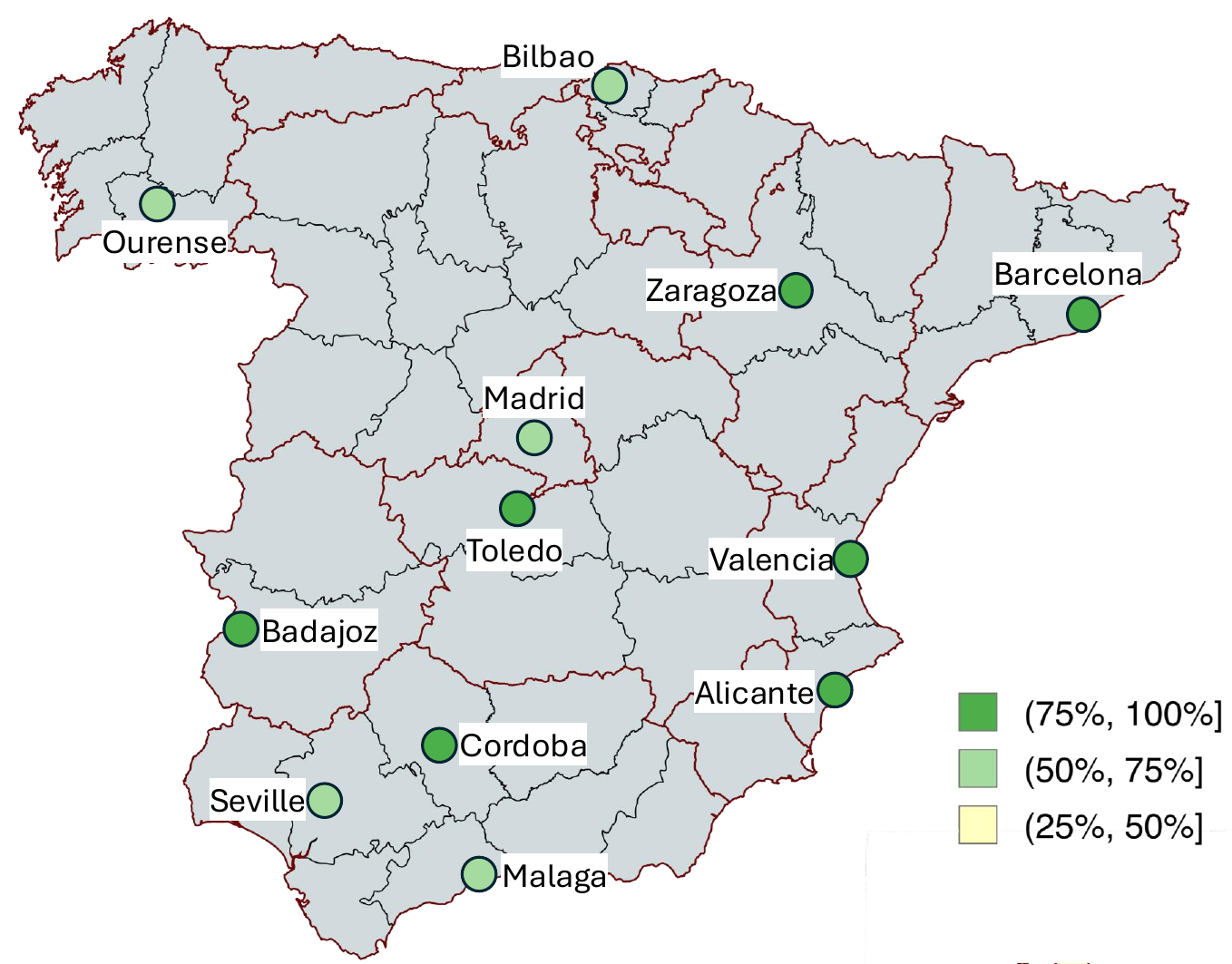}} \hfill
    \subfloat[Map chart of city-level prediction accuracies on L2 heatwaves]{\includegraphics[width = 0.47\linewidth]{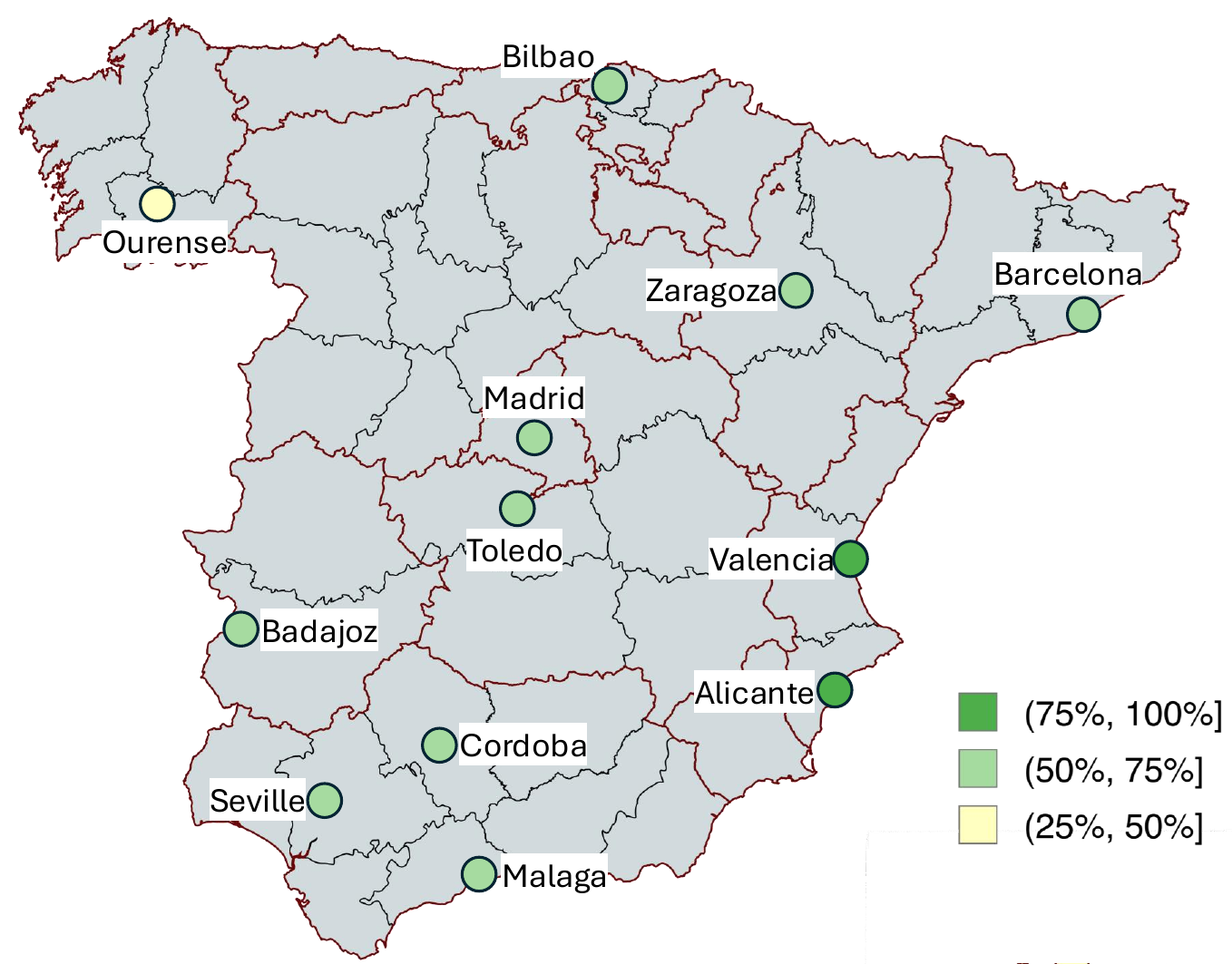}} \\
    \caption{\textbf{A visualization of \model's performance on Spain (mainland).} Figure (a) (b) shows the map chart noting the provincial prediction accuracies, and Figure (c) (d) notes the city-level prediction accuracies. Provinces outside Iberian Peninsula are hidden for brevity.}
    \label{fig:heatmap}
\end{figure}


To demonstrate the overall performance of \model in each city and province, we show the evaluation results aggregated over the entire study period in Table \ref{fig:spatialResult}, along with the performance of baseline methods for comparison. The results are shown in four metrics: 1) Accuracy, 2) Precision, 3) Recall, and 4) F1-score. We show the definition of metrics in Supplementary. Intuitively, higher precision indicates fewer false alarms (i.e., fewer false positives);
Higher recall reflects fewer missed alarms (i.e., fewer false negatives);
Higher accuracy signifies better overall correctness in labeling heatwaves (deadly or not);
Higher F1-score suggests a balanced reduction in both false and missed alarms.
Additionally, Figure \ref{fig:heatmap} visualizes the spatial distribution of prediction accuracies.

For provincial L1 heatwave prediction (Table \ref{subfig:momoTable}), \model achieved an average accuracy of $77.0\%$, with $80.6\%$ precision, $87.4\%$ recall, and $83.8\%$ F1-score. These results indicate that, over 17 out of 20 issued alarms corresponded to actual L1 heatwaves, over 16 out of 20 L1 heatwaves were detected, and over 15 out of 20 model predictions were correct. Performance for L2 heatwave prediction was slightly lower, with $66.4\%$ accuracy, $55.6\%$ precision, $64.8\%$ recall, and $59.8\%$ F1-score. These results indicate that, approximately 11 out of 20 issued alarms corresponded to actual L2 heatwaves, around 13 out of 20 L2 heatwaves were detected, and roughly 13 out of 20 predictions were correct. 


Notably, in provinces experiencing more than three heatwaves annually (e.g., Madrid, Seville, Zaragoza, Córdoba, and Badajoz), \model achieved higher performance for L1 heatwave prediction (Table \ref{subfig:momoTable}), with an average precision of $84.6\%$ and recall of $89.0\%$, surpassing results in other provinces ($81.6\%$ precision, $87.4\%$ recall). This disparity was even more pronounced for L2 heatwave prediction: high-frequency provinces attained $60.9\%$ precision and $66.4\%$ recall, while others showed lower precision ($44.4\%$) but higher recall ($75.0\%$). Yet, this trend is not observed in city-level predictions (Table \ref{subfig:ineTable}), where most of the cities are equipped with plenty of heatwave records. In the deadly prediction on the five aforementioned cities, \model received $80.4\%$ precision and $89.5\%$ recall, close to the other cities ($81.5\%$ precision, $88.2\%$ recall). Similarly, in L2 heatwave predictions, \model received $66.7\%$ precision and $70.0\%$ recall, which is lower than the other cities ($68.7\%$ precision, $73.0\%$ recall). 


\subsection{Temporal Robustness: Performance across Different Years}

\begin{figure}[!t]
    \centering
    \resizebox{\textwidth}{!}{%
    \begin{tikzpicture}
    \node[inner sep=5pt] (continentalContent) {
        \begin{tabular}{ccc}
            \hspace{-0cm}
            \subfloat[Badajoz]{\includegraphics[width=0.3\linewidth]{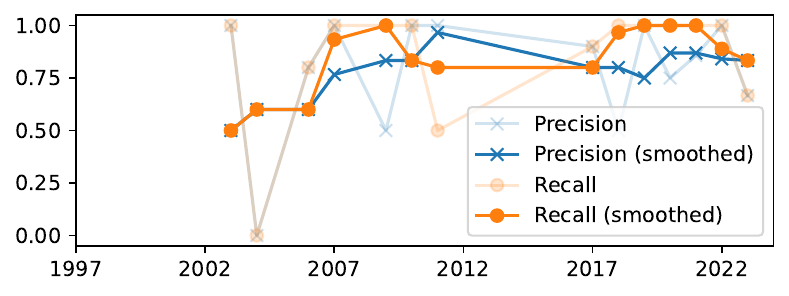}} &
            \subfloat[Cordoba]{\includegraphics[width=0.3\linewidth]{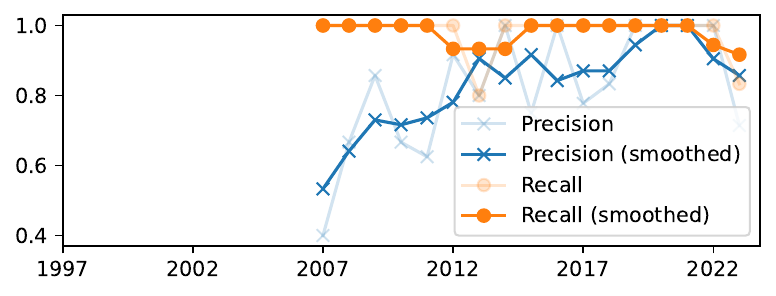}} &
            \subfloat[Madrid]{\includegraphics[width=0.3\linewidth]{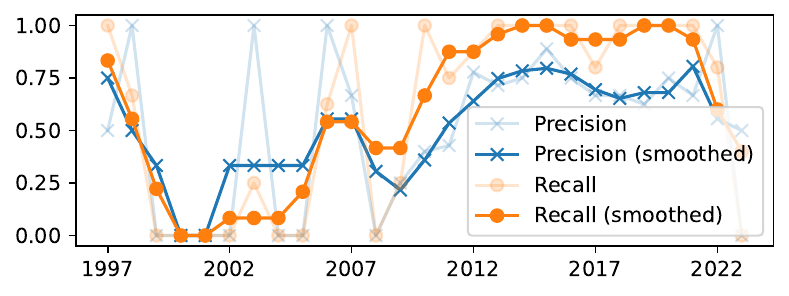}} \\
            \subfloat[Seville]{\includegraphics[width=0.3\linewidth]{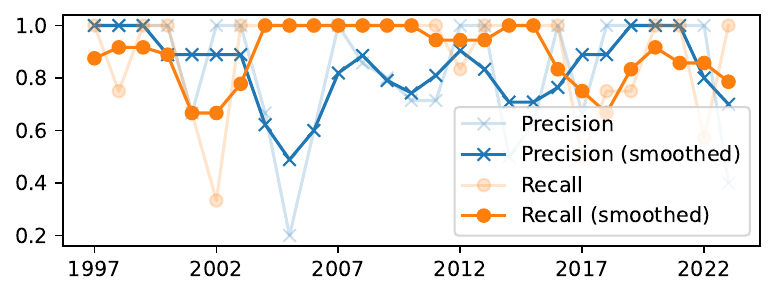}} &
            \subfloat[Toledo]{\includegraphics[width=0.3\linewidth]{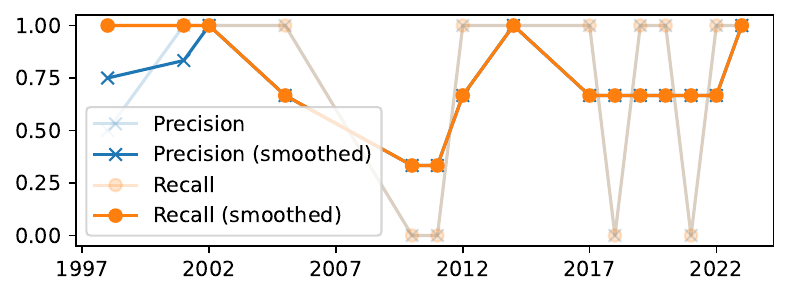}} &
            \subfloat[Zaragoza]{\includegraphics[width=0.3\linewidth]{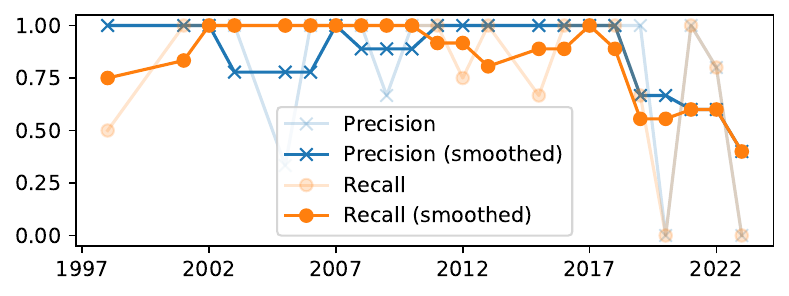}} \\
        \end{tabular}
    };

    \node[draw, solid, inner sep=5pt, below left=0.9cm and 0.5cm of continentalContent.south west, anchor=north west] (mediterranean) {
        \begin{tabular}{cc}
            \subfloat[Alicante]{\includegraphics[width=0.3\linewidth]{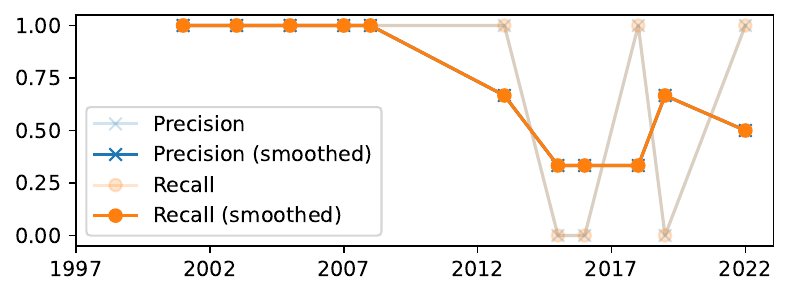}} &
            \subfloat[Barcelona]{\includegraphics[width=0.3\linewidth]{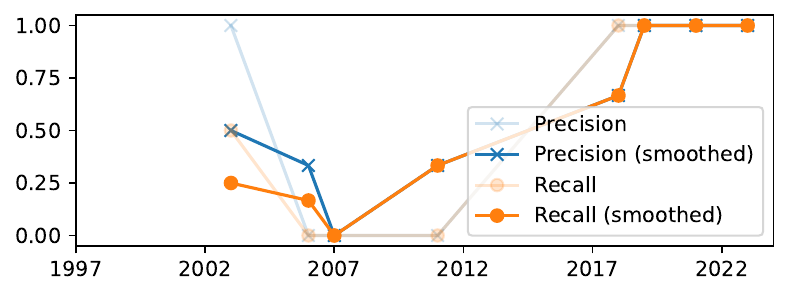}} \\
            \subfloat[Malaga]{\includegraphics[width=0.3\linewidth]{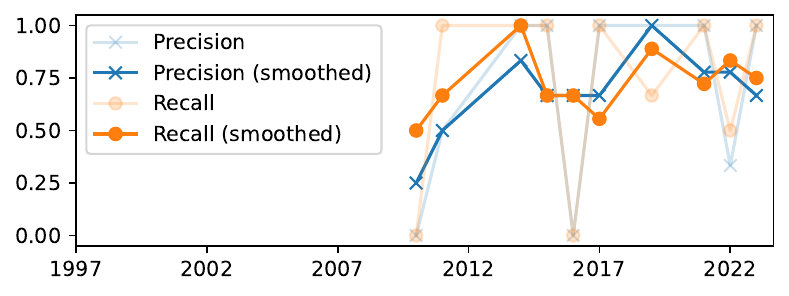}} &
            \subfloat[Valencia]{\includegraphics[width=0.3\linewidth]{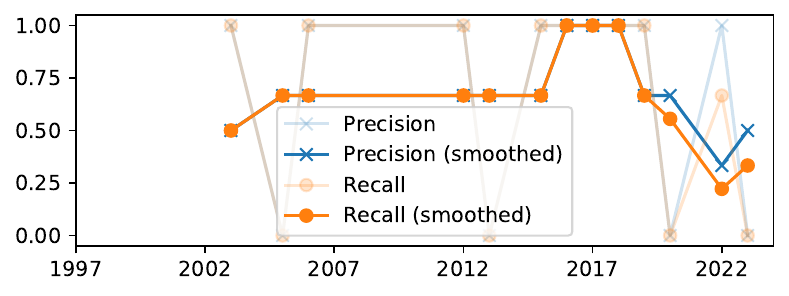}} 
        \end{tabular}
    };

    \node[draw, solid, inner sep=5pt, below right=0.9cm and 0.5cm of continentalContent.south east, anchor=north east] (atlantic) {
        \begin{tabular}{c}
            \subfloat[Bilbao(Biscay)]{\includegraphics[width=0.3\linewidth]{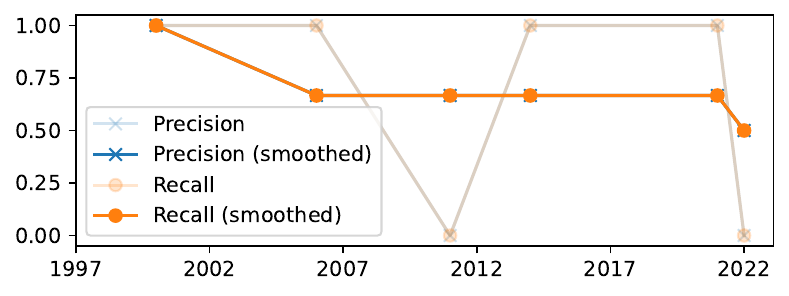}} \\
            \subfloat[Orense]{\includegraphics[width=0.3\linewidth]{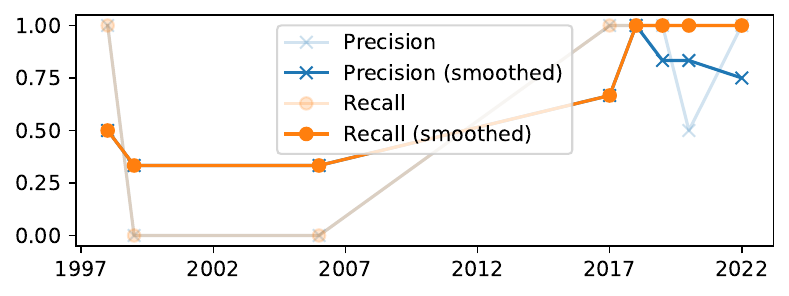}} \\
        \end{tabular}
    };

    \node[fit=(mediterranean)(atlantic), inner sep=0pt] (bottomgroup) {};

    \node[draw, solid, inner sep=5pt, minimum width=18.6cm, minimum height=6cm, anchor=south, above=1cm of continentalContent.north] (continental) at (bottomgroup.north) {\phantom{X}};

    \node[below=0.2cm of continental.south] {\textbf{Continental Area}};
    \node[below=0.2cm of mediterranean.south] {\textbf{Mediterranean Area}};
    \node[below=0.2cm of atlantic.south] {\textbf{Atlantic Area}};
    \end{tikzpicture}}

    \caption{\textbf{\model demonstrates consistent performance across most years, with improved accuracy as more data becomes available. }
    The figures present annual heatwave prediction results (smoothed using a three-year sliding window) for city-level and provincial data from 1997 to 2023 across 12 selected Spanish cities, evaluated by Precision and Recall metrics. Cities are grouped by climate zone: 1) Continental, 2) Mediterranean, and 3) Atlantic.
    \model maintains stable performance across most years while showing gradual improvement over time, except during the COVID-19 pandemic period. Notably, despite the significant challenges of mortality prediction during the pandemic, \model achieves strong performance in recent years, particularly in Continental regions with more frequent heatwave occurrences. This suggests the model's particular effectiveness in areas with higher heatwave prevalence, highlighting its potential for broad impact in vulnerable regions. 
    }
    \label{fig:yearlyAblation}
\end{figure}

To demonstrate the temporal variance of \model's performance, we summarize the precision and recall of city-level L1 heatwave prediction results year-by-year. As shown in Figure \ref{fig:yearlyAblation}, the model receives 100 percent precision and recall (meaning all heatwaves are correctly classified) on more recent years, but the performance falls more in earlier years, especially before 2005. Among all years, the prediction result in 2016 is the best, with all 7 cities with heatwaves receiving over $80\%$ precision and recall. The worst prediction result comes from the years 2020 and 2021, where approximately 3 out of 10 cities with heatwaves received precision and recall lower than $70\%$. An intuitive answer to this drop might be the occurrence of the COVID-19 pandemic, which we discuss in more detail later. It is also worth mentioning that the trends in precision and recall over time are stable in the Continental area, but not as stable in the Mediterranean and Atlantic areas, where there is no heatwave occurrence over years.  

\subsection{Demographic Generalizability: Age-Specific Performance}

\begin{figure}[!t]
    \centering
    \subfloat[City-level Precisions]{\includegraphics[width = 0.24\linewidth]{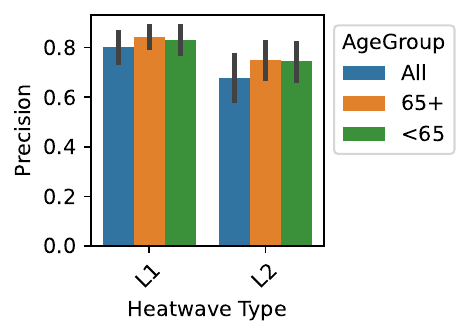}\label{subfig:ageINEdeadly}}
    \hfill
    \subfloat[City-level Recalls]{\includegraphics[width = 0.24\linewidth]{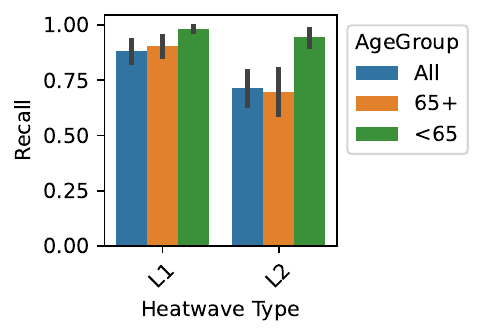}\label{subfig:ageINEextreme}}
    \hfill
    \subfloat[Provincial Precisions]{\includegraphics[width = 0.24\linewidth]{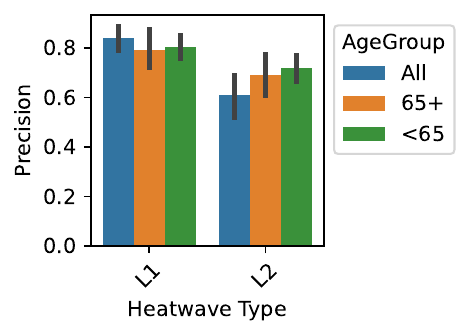}\label{subfig:ageMOMOdeadly}}
    \hfill
    \subfloat[Provincial Recalls]{\includegraphics[width = 0.24\linewidth]{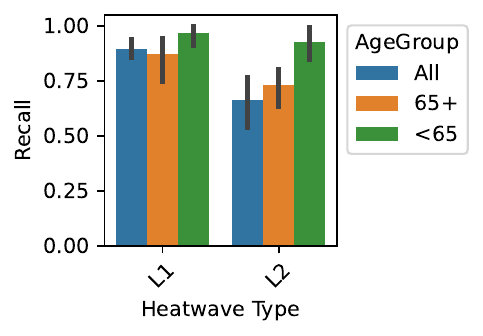}\label{subfig:ageMOMOextreme}}
    \\
    \subfloat[City-Level precision change from All to 65+]{
    \includegraphics[width = 0.49\linewidth]{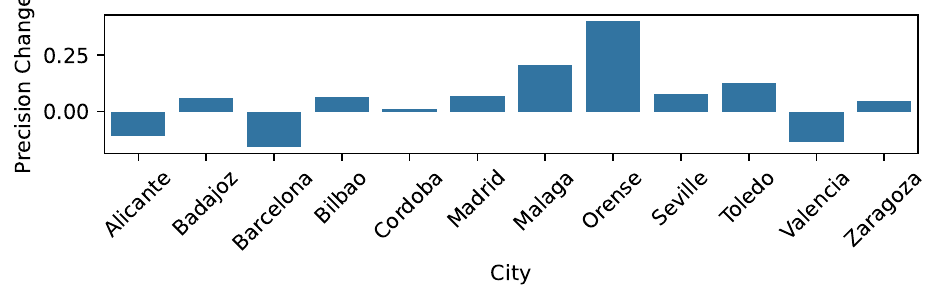}\label{subfig:INEpr65+Change}}
    \hfill
    \subfloat[City-Level recall change from All to 65+]{
    \includegraphics[width = 0.49\linewidth]{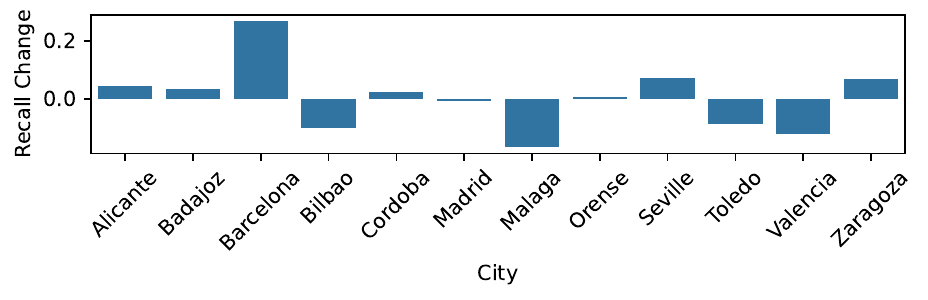}\label{subfig:INErc65+Change}}
    \\
    \subfloat[City-level precision change from All to Under-65]{
    \includegraphics[width = 0.49\linewidth]{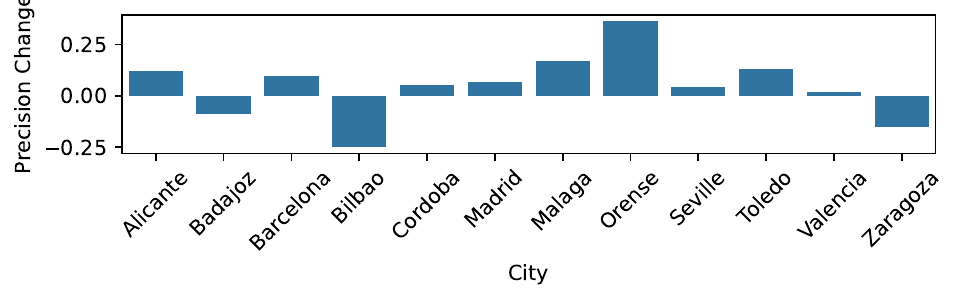}\label{subfig:INEprunder-65Change}}
    \hfill
    \subfloat[City-level recall change from All to Under-65]{
    \includegraphics[width = 0.49\linewidth]{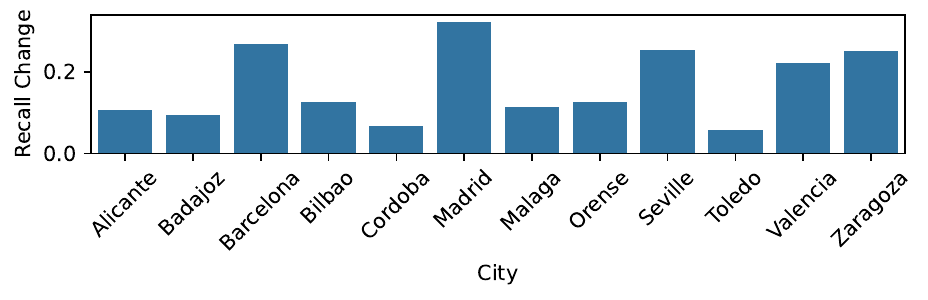}\label{subfig:INErcunder-65Change}} 
    \\
    \subfloat[Provincial precision change from All to 65+]{
    \includegraphics[width = 0.24\linewidth]{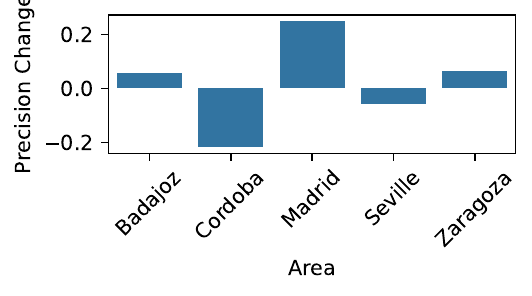}\label{subfig:MOMOpr65+Change}}
    \hfill
    \subfloat[Provincial recall change from All to 65+]{
    \includegraphics[width = 0.24\linewidth]{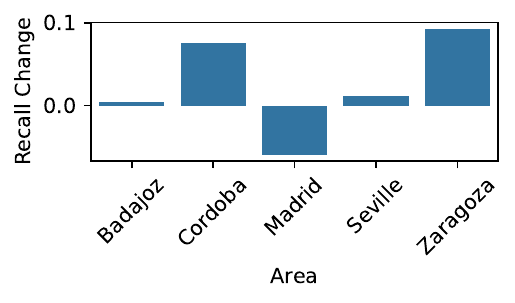}\label{subfig:MOMOrc65+Change}}
    \hfill
    \subfloat[Provincial precision change from All to Under-65]{
    \includegraphics[width = 0.24\linewidth]{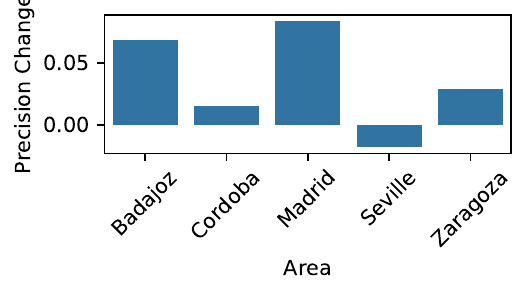}\label{subfig:MOMOprunder-65Change}}
    \hfill
    \subfloat[Provincial recall change from All to Under-65]{
    \includegraphics[width = 0.24\linewidth]{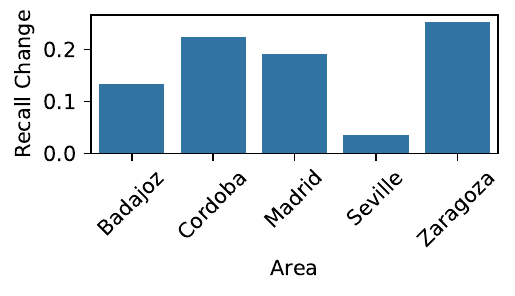}\label{subfig:MOMOrcunder-65Change}}
    \\
    \caption{\textbf{\model demonstrates consistent performance across both younger (under 65) and older (65+) population groups in all cities.} We evaluated \model by replacing all-age mortality records with age-specific mortality data (Under-65 and 65+ populations). Figures (a)-(d) present the average precision and recall for both age groups across all cities and regions, while (e)-(l) display city-level and provincial variations in performance metrics (averaged across L1 and L2 heatwaves) compared to all-age population predictions. The results show that \model maintains comparable performance when predicting heatwave impacts for both age groups relative to all-age predictions. 
    }
    \label{fig:ageAblation}
\end{figure}

As different ages of population may be affected by heatwaves in different ways, we demonstrate the performance of \model on different ages by three evaluation runs: 1) All ages, 2) Under-65 years of age, 3) 65+ years of age. Specifically, across these three runs, we use the same training and prediction pipeline as well as non-mortality data, but update the all-cause mortality data accordingly. Figure \ref{fig:ageAblation} shows the results for each age group stratified by type of heatwave and city. 

Generally, on city-level predictions, the performance of \model in the 65+ and Under-65 population is similar to that of the entire population, except for the L2 heat waves, where \model yields a much higher recall in the Under-65 population (Figure \ref{subfig:ageINEdeadly}, \ref{subfig:ageINEextreme}). For 65+ predictions, compared to the results on all-age, 65+ predictions yield higher precision in 9/12 cities with $12.6\%$ average absolute difference to all-age (Figure \ref{subfig:INEpr65+Change}), and higher recall in 7/12 cities with $10.7\%$ average absolute difference (Figure \ref{subfig:INErc65+Change}). These lead to cumulatively $5.5\%$ higher precision and $0.3\%$ higher recall. For Under-65 predictions, 9 cities yield higher precision with $13.0\%$ average absolute difference to all-age, and all cities yield higher recall with $16.7\%$ average absolute difference. These lead to cumulatively $4.7\%$ higher precision and $16.7\%$ higher recall.

Same studies are also conducted in provincial predictions. To ensure enough heatwaves for each population, we only study the 5 cities with most heatwaves. Generally, \model performs close on L1 heatwave predictions, but gets higher precision and recall in L2 heatwave predictions (Figure \ref{subfig:ageMOMOdeadly}, \ref{subfig:ageMOMOextreme}). For 65+ predictions, compared with all-age results, \model receives higher precision in 2 cities with $5.8\%$ mean absolute difference (Figure \ref{subfig:MOMOpr65+Change}), and higher recall in 3 cities with $3.4\%$ mean absolute difference (Figure \ref{subfig:MOMOrc65+Change}). These lead to cumulatively $0.8\%$ higher precision and $1.0\%$ higher recall. For Under-65 predictions, \model receives higher precisions for 4 cities with $3.5\%$ absolute difference, and higher recall for all cities with $7.0\%$ absolute difference. This leads to cumulatively $1.5\%$ higher precision and $7.0\%$ higher recall.

\subsection{Predictive Synthesis: Advantage of Utilizing Historical Mortality Data}

\begin{figure}[!t]
    \centering
    \subfloat[Precision and Recall of provincial predictions against baseline models]{\includegraphics[width = 0.24\linewidth]{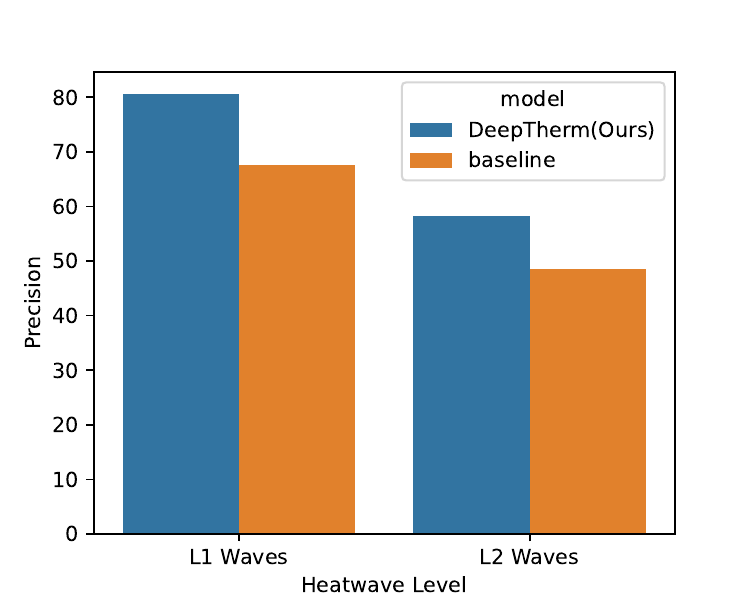}
    \includegraphics[width = 0.24\linewidth]{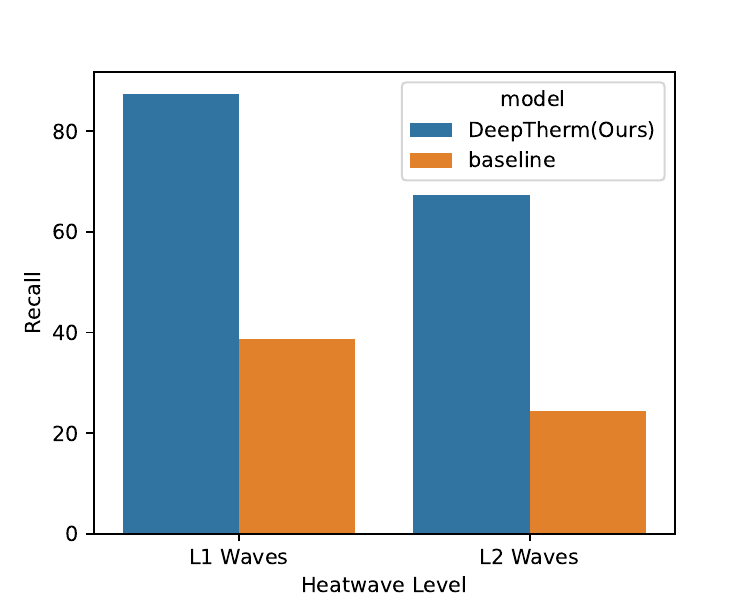}}
    \hfill
    \subfloat[Precision and Recall of city-level predictions against baseline models]{\includegraphics[width = 0.24\linewidth]{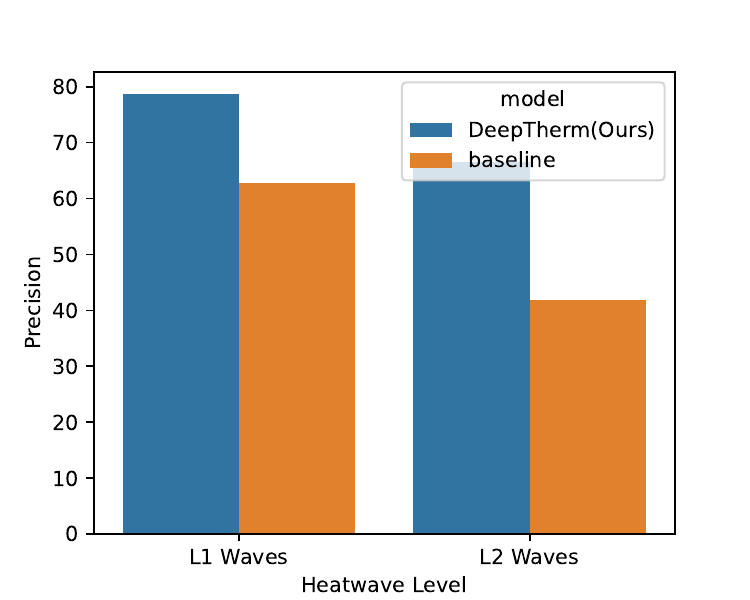}
    \includegraphics[width = 0.24\linewidth]{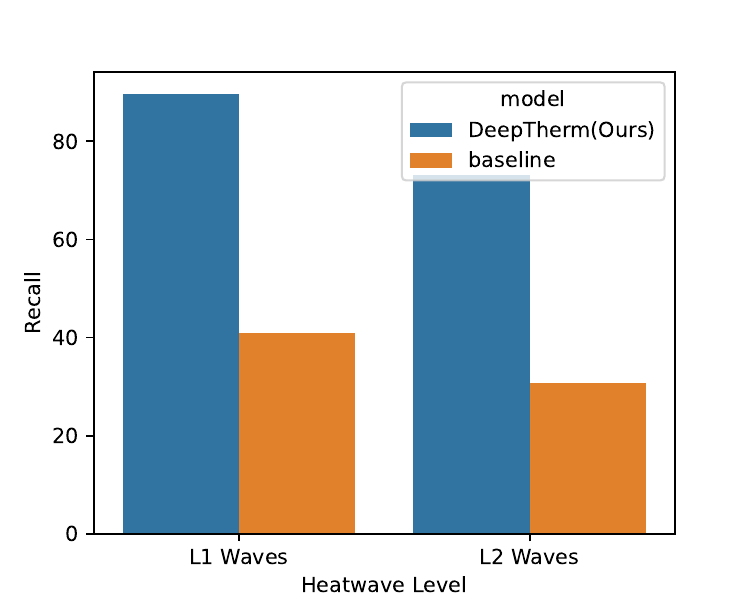}}
    \caption{\textbf{Comparing the prediction results of \model to previous deadly heatwave prediction methods that project non-mortality variables to excess mortality ratio and cannot use historical all-cause mortality records.} \model achieves significantly better recall while maintaining superior precision, implying \model predicted more missing alarms without causing more false alarms. This highlights the advantage of capability inside using historical mortality data, which is introduced by our novelly designed dual-prediction pipeline.}
    \label{fig:baseline}
\end{figure}

One key advantage of \model, compared to previous deadly heatwave prediction approaches \cite{mistry2024real}, is the capability of utilizing daily all-cause mortality history in the prediction. We examined such an advantage by comparing the predictive performance of \model with existing deadly heatwave models that leverage predictive accuracy yet are not capable of utilizing daily historical mortality \cite{mathes2017real, mistry2024real}. Table \ref{fig:baseline} shows that \ model achieved significantly higher recall while maintaining superior precision, implying that \model, compared to baseline methods, significantly cuts off the number of missing alarms without causing more false alarms. 

\subsection{Adaptive Thresholding: Balancing False Alarms and Missed Detections} \label{subsec:balancing}

\begin{figure}[!t]
    \centering
    \subfloat[False positive and false negative rates of L1 predictions under different decision thresholds]{\includegraphics[width = 0.5\linewidth]{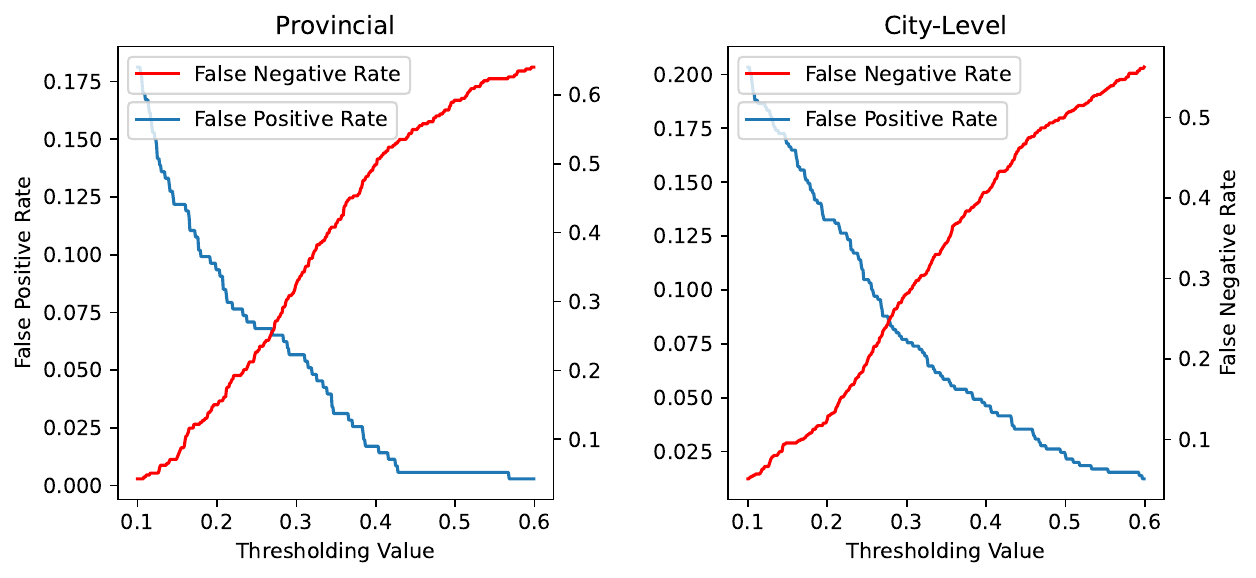}} \hfill
    \subfloat[False positive and false negative rates of L2 predictions under different decision thresholds]{\includegraphics[width = 0.5\linewidth]{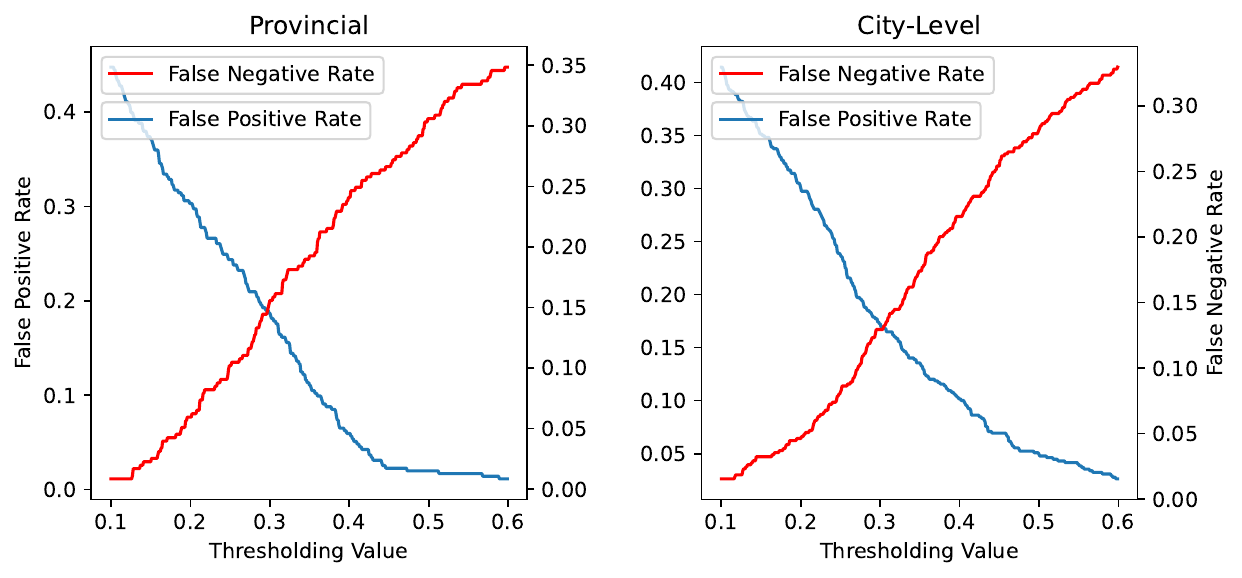}}
    \caption{\textbf{\model offers flexible threshold adjustment to balance false positives (false alarms) and false negatives (missed alarms) according to operational needs.} The curves illustrate the trade-off between false positive and false negative rates across different threshold values, demonstrating how \model can be adjusted to prioritize either alarm certainty or detection sensitivity. For provincial classifications, the ranges $[0.2, 0.25]$ and $[0.3,0.35]$ serve as robust threshold regimes for L1 and L2, respectively, as they lie between two intervals of sharp fluctuations in both false positive and false negative rates, offering a stable balance. In contrast, city-level classification rates exhibit a smoother trend, allowing greater flexibility in threshold selection. While alternative values may be chosen under stricter constraints on either rate, these adjustments require no modifications to the underlying prediction modules (all-cause and baseline) and can be applied to any region. This adaptability enables policymakers to tailor \model to diverse operational needs—whether minimizing false alarms or maximizing heatwave detection coverage.
    }
    \label{fig:balancing}
\end{figure}

While prior predictions assume fixed threshold values in \model (aligned with the Level-1 and Level-2 heatwave definitions in Section \ref{subsec:problemSetup}), \model can adapt its thresholding strategy without retraining the model parameters. This allows the heat-related mortality predictions to be calibrated for higher (or lower) sensitivity, enabling adjustable alarm thresholds.
To evaluate \model's performance under different thresholding values, we vary the threshold from $0.01$ to $0.50$ in increments of $0.001$ and plot the resulting false positive-to-false negative trade-off in Figure \ref{fig:balancing}. For provincial classifications, the model exhibits sharp trade-offs between false negative and false positive rates across most thresholds, except within the ranges $[0.2, 0.25]$ (L1) and $[0.3, 0.35]$ (L2), which emerge as robust regimes for threshold selection. In contrast, city-level classifications show smoother trade-off curves, permitting greater flexibility in threshold choice. Alternative values may also be selected if stricter constraints are imposed on either rate.

%% file: sections/discussion.tex

\section{Discussion}

Establishing early warning systems for deadly heatwaves is critical for mitigating heat-related health impacts and reducing excess mortality \cite{michelozzi2010surveillance, hess2023public}, which is yet hard to achieve due to the difficulty in defining and estimating heat-related mortality records. Existing research primarily either remains on qualitative analyses of historical correlations between heat-related mortality and environmental indicators\cite{kalkstein1997evaluation, tan2007heat, liang2022identifying, kalkstein2022increasing}, or is not capable of using the historical mortality information in the prediction of near-future excess heat-related mortality \cite{mathes2017real, mistry2024real}. In contrast, \model provides a novel dual-line prediction pipeline, which enables operation using all-cause mortality records, significantly expanding its potential deployment to regions with high-quality surveillance data on daily mortality. 

Our comprehensive evaluation across Spain demonstrates the model's robust performance, indicating great potential in real-world scenarios. First, spatial evaluation shows that \model correctly identifies over $75\%$ of L1 heatwaves and over half of L2 heatwaves correctly (Table \ref{subfig:momoTable}, \ref{subfig:ineTable}), with the latter representing a particularly challenging prediction task due to their stricter excess mortality thresholds. Second, temporal evaluation of \model reveals consistent performance across years(\ref{fig:yearlyAblation}), with only minor impacts from anomalous events like the COVID-19 pandemic. Third, \model also successfully adapts to different age groups, accurately capturing variations in heatwave mortality patterns across demographics (\ref{fig:ageAblation}). 

Notably, \model achieved the aforementioned performance without requiring surveillance data that specifically identifies individual deaths as being "heat-related", relying instead only on all-cause mortality data. This is a significant advantage for real-world deployment, as determining which deaths are attributable to heat is complex and tends to underestimate the total burden of heat on mortality. Specifically, counting deaths attributed as "heat-related" underestimates total excess deaths attributable to heat by about an order of magnitude \cite{weinberger2020estimating, shindell2020effects}. We distinguish such an advantage by the superior performance of \model against baseline deadly heatwave prediction methods that are not capable of using daily all-cause mortality data as \model.

Furthermore, \model exhibits improved performance with more years of data as well as more heatwaves. 
For example, \model achieves higher accuracy in provinces and cities with more extensive heatwave records (Figure \ref{fig:spatialResult}) and in more recent years (Figure \ref{fig:yearlyAblation}). This positive correlation between data volume and model performance suggests \model is particularly well-suited for regions experiencing frequent heatwaves, where an effective early warning system could yield the most significant public health benefits by preventing heat-related mortality.

Another key strength of \model lies in its ability to effectively balance false alarms (i.e., false positives) and missed detections (i.e., false negatives) - a critical feature for real-world implementation. As heatwave alerts often directly inform policy decisions or activation of emergency or other community responses \cite{lowe2011heatwave, kravchenko2013minimization}, minimizing false alarms is essential to prevent unnecessary resource allocation while maintaining detection sensitivity. \model addresses this need through adjustable classification thresholds (Figure \ref{fig:balancing}), enabling emergency managers to optimize the trade-off between false positive and false negative rates. Importantly, the model maintains a smooth balance between these error types, avoiding disproportionate decrease in either metric during threshold adjustment.

Still, the current implementation of \model faces several constraints. 
First, \model is currently designed to use real-time non-mortality and mortality data. Meanwhile, its effectiveness in regions with reporting delays—such as those serviced by the U.S. CDC , where mortality data may lag by months (\cite{hoyert2025maternal})—remains uncertain.
Additionally, \model's robustness to quality issues of collected data—such as missing or inaccurate temperature records—remains unverified. Furthermore, our validation has been conducted exclusively in Spain, where infrastructure and climate adaptation resources are relatively homogeneous across regions. Future research should evaluate \model’s performance in more diverse spatial contexts, particularly in areas with varying levels of heat resilience infrastructure, with ablations of various input availability.

%% file: sections/method.tex
\section{Materials and Methods}

\subsection{Evaluation Data} \label{subsec:dataset}

We collected three forms of data that is required to evaluate \model: non-mortality data, all-cause mortality data, and synoptic weather-typing data. 

\noindent \textbf{Non-mortality Data} is used as input for the all-cause mortality prediction. We collect daily meteorological data from Aemet Opendata \footnote{https://www.aemet.es/en/datos\_abiertos/AEMET\_OpenData} as our non-mortality data, which comes from stations that monitor weather at local airports, where available, else (Orense, Toledo) downtown weather stations. The collected data including four variables: temperature (Celsius), pressure (hPa), wind speed (m/s), and humidity (percentage). 

\noindent \textbf{All-cause Mortality Data} is used in both all-cause and baseline mortality prediction. We collect 1) city-level daily mortality data from 1995 to 2023, released by Instituto Nacional de Estadística \footnote{https://www.ine.es/en/} (INE) 2) provincial daily mortality data from 2015 to 2023, collected by MoMo \footnote{https://momo.isciii.es/panel\_momo/}. Specifically, INE data is a private dataset where our data was obtained upon request; MoMo is a publicly available dataset, where we get the data through the public panel. 

\noindent \textbf{Synoptic Weather-typing Data} is used to predict the occurrences of heatwaves in near-future. We collect the data from the online panel of Spatial Synoptic Classification v3.0 \footnote{https://sheridan.geog.kent.edu/ssc3.html}, which contains the SSC codes for most of the main cities worldwide.

We collected data from twelve provinces and their corresponding homonymous capital cities, except Biscay (province), whose capital city is Bilbao. During the selection of cities and provinces, population was a major criterion. The final set includes nine of the twelve most populated cities in Spain, including the top six. We also considered a variety of climatic scenarios as shown in Figure \ref{fig:areaOverview}. Specifically, we balanced coast cities (4), interior cities (6), and interior cities close to the mouth of a river (2, Seville and Bilbao). The provincial set includes the six most populous provinces in Spain, accounting for some $45\%$ of the population of the country. Seven provinces are landlocked. Orense is a low-populated, landlocked province in Galicia (Northwestern Spain, Atlantic region). However, the city of Orense has a hot summer Mediterranean climate (Csb, in the K\"oppen-Geiger
climate classification) associated with the orography and served as a test for our model's performance in such conditions. An overview of the population sizes and appearance of the heatwaves of each city and the corresponding province is shown in Figure \ref{fig:areaOverview}, where the population data are collected from the NUTS3 provincial population database and Geonames city population database.



\begin{table}[t]
    \centering

    \centering
    \subfloat[Population and Climate Overview]{
    \resizebox{\linewidth}{!}{
    \begin{tabular}{c|cccccccccccc}
    \toprule
    & Alicante & Badajoz & Barcelona & Bilbao(Biscay) & Cordoba & Madrid & Malaga & Orense & Seville & Toledo & Valencia & Zaragoza \\ \midrule
    \begin{tabular}[c]{@{}c@{}}Provincial\\ Population(k)\end{tabular} & 1993 & 666 & 5878 & 1159 & 774 & 7009 & 1775 & 305 & 1969 & 743 & 2711 & 988 \\ \midrule
    \begin{tabular}[c]{@{}c@{}}City-level\\ Population(k)\end{tabular} & 335 & 148 & 1620 & 346 & 326 & 3256 & 578 & 105 & 684 & 84 & 792 & 675 \\ \midrule
    Climate (city) & Bsh & Csa & Csa & Cfb & Csa & Bsk/Csa & Csa & Csb & Csa & Bsk & Bsh & Bsk \\ \bottomrule
    \end{tabular}
    }
    }
    
    \subfloat[Heatwave Statics]{\resizebox{\linewidth}{!}{\begin{tabular}{c|c|cccccccccccc|c}
\toprule
 &  & Alicante & Badajoz & Barcelona & Bilbao & Cordoba & Madrid & Malaga & Orense & Seville & Toledo & Valencia & Zaragoza & \textbf{Total} \\ \midrule
\multirow{3}{*}{\begin{tabular}[c]{@{}c@{}}MoMo \\ (Provincial)\end{tabular}} & All & 3 & 49 & 4 & 2 & 53 & 50 & 13 & 6 & 36 & 11 & 9 & 19 & \textbf{255} \\
 & L1 & 1 & 40 & 3 & 1 & 39 & 32 & 8 & 5 & 26 & 10 & 6 & 14 & \textbf{185} \\
 & L2 & 0 & 26 & 1 & 1 & 25 & 14 & 1 & 4 & 19 & 4 & 4 & 11 & \textbf{110} \\ \midrule
\multirow{3}{*}{\begin{tabular}[c]{@{}c@{}}INE \\ (city-level)\end{tabular}} & All & 14 & 69 & 9 & 6 & 124 & 166 & 21 & 10 & 132 & 25 & 18 & 55 & \textbf{649} \\
 & L1 & 11 & 57 & 7 & 4 & 97 & 98 & 13 & 7 & 97 & 22 & 14 & 44 & \textbf{471} \\
 & L2 & 9 & 53 & 4 & 4 & 77 & 34 & 7 & 4 & 62 & 15 & 7 & 32 & \textbf{308} \\ \bottomrule
\end{tabular}}\label{subfig:mortLabelOverview}}\\
    \caption{\textbf{Our selected cities and provinces cover a variety of population size as well as heatwave appearance frequency.} Table (a) shows the population sizes of the selected cities in our dataset and their located provinces, as well as their climate conditions. Population numbers were retrieved from the Eurostat population by NUTS3 (Nomenclature of Territorial Units for Statistics level 3 available at \url{https://doi.org/10.2908/DEMO\_R\_PJANGRP3}) (province) and \texttt{geonames} database (\url{https://geonames.org}) (city). Codes are provided in Supplementary. Table (b) shows the count of heatwaves in each city (province) across all the collected time periods. Climate K\"oppen-Geiger codes for the period 1991-2020 were retrived from Figure~5 in Ref.~\citeonline{Chazarra2022}.  
    }
    \label{fig:areaOverview}
\end{table}

\subsection{Problem Setup}\label{subsec:problemSetup}

We formulate the deadly heatwave prediction problem correspondingly: for a certain day $t$, we have the historical all-cause mortality data $X_{:t}$, the historical non-mortality data $Z_{:t}$, and the synoptic weather-typing data extended to $h$-day future $N_{:t+h}$. If $N_{:t+h}$ shows there is a heatwave between day $t$ and $t+h$, we run \model to predict 1) whether the heatwave will be deadly or not, 2) whether the heatwave will be extreme or not. This will then involve two settings: 1) definition of heatwaves, 2) definition of deadly and L2 heatwaves

\noindent \textbf{Definition of Heatwaves} Following prior work in synoptic classification \cite{sheridan2002redevelopment, hondula2014ssc}, we employ the Spatial Synoptic Classification (SSC) pipeline to identify heatwave events. Specifically, each day is categorized into one of six types: Dry Polar (DP), Dry Moderate (DM), Dry Tropical (DT), Moist Polar (MP), Moist Moderate (MM), or Moist Tropical (MT). A heatwave is defined as a sequence of consecutive days where, for each day in the sequence, at least one of the following conditions holds over a three-day sliding window: 1) all three days are Dry Tropical days, or 2) the three-day window contains both Dry Tropical and Moist Tropical days.

\noindent \textbf{Definition of Heatwave Levels} To test the performance of \model under broader or stricter heatwave standards, we categorize heatwaves into three levels: 0, 1, 2, according to the ratio of how much the heat-related mortality exceeds baseline mortality during the heatwave. Specifically, for a heatwave spanning $N$ days, we compute such a ratio $R$ by:

\begin{equation}
    R = \max_{t \in N} \left(\frac{X_t - \tilde{X}_t}{\tilde{X}_t}\right),\label{eq:R}
\end{equation}
where $\tilde{X}_t$ and $X_t$ are baseline mortality and all-cause mortality of day $t$, respectively. While $X_t$ directly comes from data, $\tilde{X}_t$ is generated by a Quasi-Poisson regression model \cite{lin2024daily}. We establish such a Quasi-Poisson model in the same way as the baseline prediction module of \model, which will be described in detail in Section \ref{subsubsec:poissonModule}. For a heatwave, if the ratio $R$ goes beyond $15\%$, we mark the heatwave as a level 1 (L1) heatwave. And if the ratio goes beyond $30\%$, we further mark the heatwave as a level 2 (L2) heatwave. By this definition, L2 heatwaves are also classified as deadly. An overview of deadly and L2 heatwave distributions across cities and provinces is provided in Table \ref{subfig:mortLabelOverview}.

\subsection{Design of \model}

\model is composed of three components: 1) all-cause mortality prediction module 2) baseline mortality prediction module 3) decision module. 

\subsubsection{All-cause Mortality Prediction Module}

The all-cause mortality prediction module is established on a deep-learning-based predictor, named Transformer \cite{vaswani2017attention}, which produces sequence-to-sequence predictions based on the self-attention mechanism. Specifically, note the input mortality history as $X \in \mathbb{R}^{1\times T}$ where $T$ is the length of mortality history, and the non-mortality history as $Z \in \mathbf{\mathbb{R}^{m \times T}}$ where $m$ is the number of collected non-mortality variables. The input to the model will then by $X' = \{X; Z\} \in \mathbb{R}^{(m+1) \times T}$. Transformer will first produce three matrices named query $Q$, key $K$, and value $V$ depending on learnable weights $W_K, W_Q, W_V \in \mathbb{R}^{(m+1)\times d_K}$, which will be calculated towards a self-attention output:

\begin{align}
    Q &= (X')^T W_Q, \label{eq:Q}\\
    K &= (X')^TW_K,\label{eq:K}\\
    V &= (X')^TW_V ,\label{eq:V}\\
    Z' &= \mathrm{softmax}\left(\frac{QK^T}{\sqrt{d_K}}\right)V.\label{eq:Z}
\end{align}

Such a procedure will happen iteratively through every layer, that is, the input $X'$ in each layer corresponds to the output $Z'$ from the previous layer. After the final layer, the output prediction $\tilde{X}$ will be produced by a linear projection $f_p$ based on the output from the last layer: $\tilde{X} = f_p(Z')$. 

In our evaluations, we set $T = 14$, inputting the 14-day history of both observed mortality and non-mortality factors to the model, and prompt the model to predict total mortality up to 5 days in the future. All the input factors are linearly normalized regarding the historical max and min values: 

\begin{equation}
\hat{X}_t = \frac{X_t - \min_{j \in [1:t)} X_{j}}{\max_{j \in [1:t]}X_j -\min_{j \in [1:t)}X_{j}}.
\end{equation}

We train the model by minimizing the mean squared error (MSE) between the prediction and the ground truth mortality. Such training is done in a \textit{real-time} manner. Specifically, when predicting the mortality for year $Y$, we assume the model is accessible to all mortality data up to year $Y-1$. Given that our data is collected from 1995 to 2023, we simulated the prediction procedure each year from 1997 to 2023, updating the model of each year based on the trained model from the previous year. Details of training and model hyperparameters can be found in Supplementary. 

\subsubsection{Baseline Mortality Prediction Module} \label{subsubsec:poissonModule}

Following the approach outlined in a previous study \cite{lin2024daily}, we estimate future baseline mortality using a Quasi-Poisson regression model. Specifically, we fit the model to historical all-cause mortality data, incorporating selected covariates that account for temporal trends, such as day-of-the-week and holiday indicators. We then apply the fitted model to the covariates from the prediction horizon to generate corresponding mortality estimates. Since the inputs to the Quasi-Poisson model exclude heat-related variables, this modeling process effectively acts as a denoising procedure, disentangling heat-related mortality from all-cause mortality. Thus, the resulting predictions can be interpreted as approximations of baseline mortality, independent of heat-related influences. Empirically, when training this baseline mortality predictor, we restrict the training data to all-cause mortality records from the past two years. This constraint is necessary because baseline mortality may exhibit distributional shifts over longer time periods due to population changes.

\subsubsection{Decision Module}

Given the predicted all-cause mortality and baseline mortality in the future, the decision module determines whether to raise an alarm for a heatwave based on weather-typing data. Specifically, for a heatwave period $T = [i, i+h]$ spanning days $i$ to day $i+h$, let $X_{all-cause}(t)$ (where $t \in [i,i+h]$) denote the predicted all-cause mortality for day $t$, and $X_{non-hr}(t)$ represent the predicted baseline mortality. The decision module first calculates the heat-related mortality by subtracting baseline mortality from all-cause mortality:

\begin{align}
    X_{hr}(t) = X_{all-cause}(t) - X_{non-hr}(t),\ \forall t \in T
\end{align}

Next, it computes the relative excess of heat-related mortality over baseline mortality for each day:
\begin{align}
    R(t) = \frac{X_{hr}}{X_{non-hr}}=\frac{X_{all-cause}(t) - X_{non-hr}(t)}{X_{non-hr}(t)}, \ \forall t \in T
\end{align}

If any $R(t)$ exceeds a predefined threshold $\alpha$, an alarm is triggered for the heatwave. Without changing the predictions of $X_{all-cause}$ and $X_{non-hr}$, we can balance the alarms of heatwaves by flexibly adjusting $\alpha$, which is investigated in Section \ref{subsec:balancing}. 

\subsection{Baseline Deadly Heatwave Prediction Models}

One key advantage of \model is the capability of using all-cause mortality history, which is not yet achieved by existing deadly heatwave prediction methods. We hereby introduce two baselines from the previous deadly heatwave prediction methods: \cite{mathes2017real} performs a linear projection from seasonal splines of time and temperatures to excess mortality. \cite{mistry2024real} performs an exponential projection from daily temperature and annual average mortality to excess mortality. We maintained their calculation of the excess mortality ratio. Furthermore, we allowed a calibration in model parameters as well as decision threshold upon the same training data as \model for a fair comparison.

%% file: sections/supplementary.tex
\section*{Supplementary Information}



\section*{Data Collection}

We show the details of the collected data here.

\subsection*{Mortality Data}

\subsubsection*{City-Level}

We collect city-level mortality data through requests to INE database \footnote{https://www.ine.es/en/}, maintained by Instituto Nacional de Estadística (National Statistics Institute) of Spain. The collected data contains daily all-cause mortality records of Hombre (Men) and Mujer (Women) that is aggregated within 5-year bins of age. We aggregate the mortality through both genders.

\subsubsection*{Provincial}
We collect provincial mortality data through the public panel of MoMo \footnote{https://momo.isciii.es/panel\_momo/}, maintained by National Institute of Health Carlos III (ISCIII). The collected data contains daily all-cause mortality records aggregated into eight age bins: “all”, “0-14”, “15-44”, “45-64”, “65-74”, “75-84”, “+65”, and “+85”. We use the \textit{defunciones\_observadas} column in data as the daily mortality, aggregating between age bins according to corresponding settings.

\subsection*{Non-mortality Data}

We collect daily non-mortality data from the Aemet Opendata \footnote{https://www.aemet.es/en/datos\_abiertos/AEMET\_OpenData}, maintained by Ministerio para la Transición Ecológica y el Reto Demográfico (Ministry for the Ecological Transition and the Demographic Challenge) of Spain. The collected data contains daily averages of temperature (Celsius), pressure (hPa), wind speed(m/s), and humidity (percentage) of near-land air, starting from 1970. 

\subsection*{Synoptic Weather-typing Data}

We collect synoptic weather-typing data from the public panel of Spatial Synoptic Classification V3.0 \footnote{https://sheridan.geog.kent.edu/ssc3.html}. The collected data contains daily synoptic weather prototypes, starting from 1970s or 1980s. Instead of obtaining the original synoptic codes in our input data, we preprocess them into boolean heatwave indicators according to the definition described in Section 4.2. 

\section*{Evaluation Metrics}

We evaluate \model under a binary classification setting: heatwaves which are labeled as deadly (extreme) or safe, and \model predicts if these heatwaves are deadly (extreme) or safe. The predictions from \model are then evaluated into one of the four classes:

\begin{itemize}
    \item[1] True Positive (TP), where \model correctly recognize a deadly (extreme) heatwave 
    \item[2] False Positive (FP), where \model wrongly categorize a safe heatwave as a deadly (extreme) heatwave. The occurrence of FP will lead to a false alarm
    \item[3] False Negative (FN), where \model wrongly categorize a deadly (extreme) heatwave as a safe heatwave
    \item[4] True Negative (TN), where \model correctly recognize a safe heatwave
\end{itemize}

Given the counted number of these four classes, we calculate the evaluation metrics correspondingly:

\begin{align}
    \mathrm{Accuracy} = \frac{\mathrm{TP} + \mathrm{TN}}{\mathrm{TP} + \mathrm{TN}+ \mathrm{FP}+ \mathrm{FN}} \\
    \mathrm{Precision} = \frac{\mathrm{TP}}{\mathrm{TP} + \mathrm{FP}} \\
    \mathrm{Recall} = \frac{\mathrm{TP}}{\mathrm{TP} + \mathrm{FN}} \\
    \mathrm{F1} = \frac{\mathrm{Recall} \times \mathrm{Precision}}{\mathrm{Recall} + \mathrm{Precision}}
\end{align}
    
\section*{Codes for NUTS and Geonames Dataset}

We show the codes for NUTS and Geonames dataset corresponded to each city and province, which we used to get the statics in Table 2.

\begin{table}[h]
    \centering
    \resizebox{\linewidth}{!}{
    \begin{tabular}{c|cccccccccccc}
        \toprule
        & Alicante & Badajoz & Barcelona & Bilbao(Biscay) & Cordoba & Madrid & Malaga & Orense & Seville & Toledo & Valencia & Zaragoza \\ \midrule
        NUTS3 code & ES521 & ES431 & ES511 & ES213 & ES613 & ES300 & ES617 & ES113 & ES618 & ES425 & ES523 & ES243 \\ \midrule
        Geonames ID & 2521978 & 2519240 & 3128760 & 3128026 & 2519240 & 3117735 & 2514256 & 3114965 & 2510911 & 2510409 & 2509954 & 3104324 \\ \midrule
    \end{tabular}
    }
    \caption{NUTS3 and ID geonames codes for each province (NUTS3) and city (ID).}
    \label{tab:cityCodes}
\end{table}

\section*{Implementation Details}

We describe details of 1) how we preprocess the input data before feeding them into \model 2) how we implement the mortality predictors and train them during the evaluations.

\subsection*{Data Preprocessing}

\noindent \textbf{Normalization} For mortality data, we retain the original values in city-level datasets, while provincial-level mortality data are normalized by dividing by one hundred. For meteorological data, each dimension is independently normalized using the historical maximum and mean values from all available data.

\noindent \textbf{Missing Data}

For days with missing mortality data (which only occurs in early 2020 and thus does not interfere with deadly heatwave prediction evaluation), we impute them using the mean value of the corresponding month (calculated from all previous years). For missing meteorological data, we fill gaps with the mean value of all preceding days in the dataset.

\noindent \textbf{Synoptic Codes of Heatwave Occurrence}

To streamline the evaluation of \model, we reuse pre-generated synoptic codes obtained from the official Spatial Synoptic Classification website\footnote{https://sheridan.geog.kent.edu/ssc3.html}. Since these codes are derived from corresponding meteorological data, their use does not introduce additional data requirements beyond those already available to \model, ensuring no risk of data requirement violations.

\subsection*{Mortality Predictors}

\subsubsection*{Transformer}

\begin{figure}[h]
    \centering
    \includegraphics[width=\linewidth]{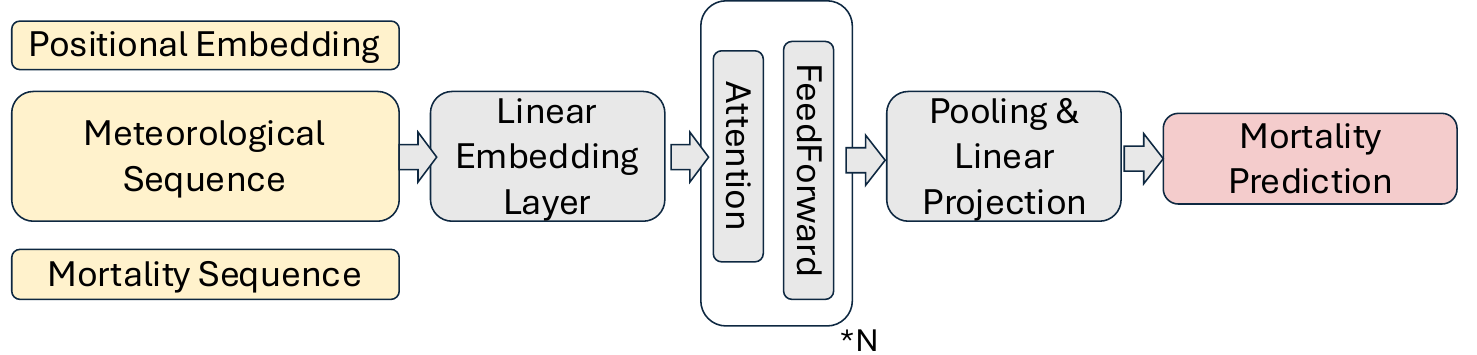}
    \caption{Overview of the structure of Transformer model, which is used to predict all-cause mortality.}
    \label{fig:transformerOverview}
\end{figure}

As illustrated in Figure \ref{fig:transformerOverview}, we employ an attention-based deep learning model, Transformer, as our all-cause mortality predictor. The model first concatenates input mortality and meteorological sequences with positional embeddings, forming a multivariate sequence where each dimension corresponds to an original feature from mortality data, meteorological data, or positional embeddings. This sequence is then projected into an embedding space via a linear transformation. The embedded input is processed recursively through an attention module to produce an output embedding sequence. Finally, the output embeddings are pooled and projected to generate the final mortality prediction. All implementations are executed using PyTorch \cite{paszke2019pytorch}.

For positional embeddings, we use a two-dimensional sequence composed of sine and cosine waves. The attention layers are configured with an embedding dimension of 32, 2 attention blocks (each with 2 attention heads), and a 2-layer MLP (hidden dimension: 32) for the final projection. In the first year of training, the model is trained for 300 epochs with a learning rate of $10^{-4}$ learning rate. For subsequent years, the model is initialized with weights from the previous year and fine-tuned for another 300 epochs at the same learning rate.

\subsubsection*{Baseline Mortality Predictor}

We use a quasi-Poisson model as our baseline mortality predictor. The predictor incorporates three input variables: 1) Day of the week (encoded as integers, where 1 = Monday and 7 = Sunday), 2) Day of the year (ranging from 0 for the first day to 364 or 365 for the last day), 3) Holiday indicator (binary, with 1 denoting an official Spanish holiday and 0 otherwise). Given these variables, we prompt the quasi-Poisson model to fit the all-cause mortality data in two years. Every time the model get re-trained, the parameters of the predictor will be calibrated from scratch.

\subsubsection*{Baseline Models}

\noindent \textbf{XGBoost} We use the implementation of XGBoost in their official python package \footnote{https://xgboost.readthedocs.io/en/release\_3.0.0/}. Specifically, we use the \textit{XGBoostRegressor}, setting the number of estimators as $1000$, learning rate as $0.01$, max depth as $6$, subsample rate as $0.8$, column sample by tree as $0.8$, and early stopping rounds as $50$. We optimize the model towards minimizing squared error, and format the input data in the same way as \model.

\noindent \textbf{Random Forest} We use the implementation of random forest from scikit-learn \cite{scikit-learn}, set the number of estimators as $1000$, and keep all other parameters as default. The input is formatted in the same way as \model.

\section*{Original Result Tables}

We show here the original experiment results of each city, each heatwave level, each population, for city-level and provincial classifications. Original prediction logs can be accessed in our Github repo. Table \ref{tab:appAllAge} shows the results of all-age classifications, Table \ref{tab:app65+} shows the results of 65+ classifications, and Table \ref{tab:appunder65} shows the results of under-65 classifications.

\begin{table}[h]
    \centering
    \begin{tabular}{l|cccc|cccc|cccc|cccc}
    \toprule
     & \multicolumn{8}{c|}{Provincial} & \multicolumn{8}{c}{City-Level} \\ \midrule
     & \multicolumn{4}{c|}{Level 1} & \multicolumn{4}{c|}{Level 2} & \multicolumn{4}{c|}{Level 1} & \multicolumn{4}{c}{Level 2} \\ \midrule
    City & TP & FP & FN & TN & TP & FP & FN & TN & TP & FP & FN & TN & TP & FP & FN & TN \\ \midrule
    Alicante & 1 & 0 & 0 & 2 & 0 & 0 & 0 & 3 & 9 & 1 & 2 & 2 & 8 & 2 & 1 & 3 \\
    Badajoz & 39 & 5 & 1 & 4 & 19 & 14 & 7 & 9 & 52 & 11 & 5 & 1 & 44 & 10 & 9 & 6 \\
    Barcelona & 2 & 1 & 1 & 0 & 0 & 1 & 1 & 2 & 5 & 0 & 2 & 2 & 2 & 1 & 2 & 4 \\
    Biscay & 1 & 1 & 0 & 0 & 0 & 0 & 1 & 1 & 4 & 2 & 0 & 0 & 3 & 0 & 1 & 2 \\
    Cordoba & 34 & 3 & 5 & 11 & 17 & 6 & 8 & 22 & 95 & 20 & 2 & 7 & 64 & 28 & 13 & 19 \\
    Madrid & 27 & 5 & 5 & 13 & 6 & 8 & 8 & 28 & 85 & 40 & 13 & 28 & 22 & 24 & 12 & 104 \\
    Malaga & 6 & 1 & 2 & 4 & 1 & 3 & 0 & 9 & 11 & 6 & 2 & 2 & 5 & 5 & 2 & 9 \\
    Orense & 5 & 1 & 0 & 0 & 3 & 1 & 1 & 1 & 7 & 3 & 0 & 0 & 3 & 5 & 1 & 1 \\
    Sevilla & 24 & 6 & 2 & 4 & 16 & 8 & 3 & 9 & 85 & 23 & 12 & 12 & 37 & 23 & 25 & 47 \\
    Toledo & 7 & 0 & 3 & 1 & 3 & 5 & 1 & 2 & 19 & 3 & 3 & 0 & 14 & 7 & 1 & 3 \\
    Valencia & 6 & 1 & 0 & 2 & 2 & 3 & 2 & 2 & 13 & 1 & 1 & 3 & 4 & 1 & 3 & 10 \\
    Zaragoza & 12 & 4 & 2 & 1 & 7 & 4 & 4 & 4 & 37 & 4 & 7 & 7 & 19 & 7 & 13 & 16 \\ \bottomrule
\end{tabular}
    \caption{The original results of all-age classifications}
    \label{tab:appAllAge}
\end{table}

\begin{table}[h]
    \centering
    \begin{tabular}{l|cccc|cccc|cccc|cccc}
    \toprule
     & \multicolumn{8}{c|}{Provincial} & \multicolumn{8}{c}{City-Level} \\ \midrule
     & \multicolumn{4}{c|}{Level 1} & \multicolumn{4}{c|}{Level 2} & \multicolumn{4}{c|}{Level 1} & \multicolumn{4}{c}{Level 2} \\ \midrule
    City & TP & FP & FN & TN & TP & FP & FN & TN & TP & FP & FN & TN & TP & FP & FN & TN \\
    Alicante & - & - & - & - & - & - & - & - & 10 & 3 & 0 & 1 & 7 & 4 & 1 & 2 \\
    Badajoz & 35 & 7 & 3 & 4 & 23 & 8 & 6 & 12 & 58 & 6 & 3 & 2 & 51 & 9 & 5 & 4 \\
    Barcelona & - & - & - & - & - & - & - & - & 6 & 2 & 0 & 1 & 3 & 2 & 1 & 3 \\
    Bilbao(Biscay) & - & - & - & - & - & - & - & - & 4 & 1 & 1 & 0 & 3 & 0 & 1 & 2 \\
    Cordoba & 34 & 15 & 2 & 2 & 19 & 17 & 6 & 11 & 100 & 18 & 3 & 3 & 72 & 30 & 9 & 13 \\
    Madrid & 20 & 1 & 12 & 17 & 9 & 2 & 8 & 31 & 82 & 33 & 23 & 28 & 27 & 20 & 27 & 92 \\
    Malaga & - & - & - & - & - & - & - & - & 13 & 3 & 4 & 1 & 6 & 2 & 7 & 6 \\
    Orense & - & - & - & - & - & - & - & - & 8 & 0 & 1 & 1 & 7 & 1 & 1 & 1 \\
    Sevilla & 21 & 8 & 1 & 6 & 10 & 6 & 2 & 18 & 98 & 14 & 6 & 14 & 52 & 24 & 25 & 31 \\
    Toledo & - & - & - & - & - & - & - & - & 22 & 1 & 1 & 1 & 14 & 3 & 7 & 1 \\
    Valencia & - & - & - & - & - & - & - & - & 12 & 2 & 1 & 3 & 3 & 2 & 6 & 7 \\
    Zaragoza & 13 & 4 & 1 & 1 & 9 & 3 & 3 & 4 & 40 & 7 & 6 & 2 & 28 & 4 & 12 & 11 \\ \bottomrule
\end{tabular}
    \caption{The original results of 65+ classifications}
    \label{tab:app65+}
\end{table}

\begin{table}[h]
    \centering
    \begin{tabular}{l|cccc|cccc|cccc|cccc}
    \toprule
     & \multicolumn{8}{c|}{Provincial} & \multicolumn{8}{c}{City-Level} \\ \midrule
     & \multicolumn{4}{c|}{Level 1} & \multicolumn{4}{c|}{Level 2} & \multicolumn{4}{c|}{Level 1} & \multicolumn{4}{c}{Level 2} \\ \midrule
    City & TP & FP & FN & TN & TP & FP & FN & TN & TP & FP & FN & TN & TP & FP & FN & TN \\
    Alicante & - & - & - & - & - & - & - & - & 13 & 1 & 0 & 0 & 13 & 1 & 0 & 0 \\
    Badajoz & 41 & 7 & 0 & 1 & 35 & 12 & 1 & 1 & 50 & 18 & 0 & 1 & 48 & 18 & 1 & 2 \\
    Barcelona & - & - & - & - & - & - & - & - & 9 & 0 & 0 & 0 & 6 & 1 & 2 & 0 \\
    Bilbao(Biscay) & - & - & - & - & - & - & - & - & 4 & 2 & 0 & 0 & 3 & 3 & 0 & 0 \\
    Cordoba & 45 & 7 & 0 & 1 & 42 & 9 & 0 & 2 & 102 & 17 & 1 & 4 & 87 & 26 & 4 & 7 \\
    Madrid & 32 & 10 & 5 & 3 & 19 & 9 & 5 & 17 & 116 & 40 & 2 & 8 & 74 & 63 & 4 & 25 \\
    Malaga & - & - & - & - & - & - & - & - & 14 & 4 & 1 & 2 & 12 & 5 & 2 & 2 \\
    Orense & - & - & - & - & - & - & - & - & 9 & 1 & 0 & 0 & 9 & 1 & 0 & 0 \\
    Sevilla & 29 & 7 & 1 & 2 & 20 & 12 & 3 & 4 & 101 & 29 & 0 & 2 & 86 & 34 & 2 & 10 \\
    Toledo & - & - & - & - & - & - & - & - & 22 & 2 & 1 & 0 & 21 & 3 & 1 & 0 \\
    Valencia & - & - & - & - & - & - & - & - & 16 & 1 & 1 & 0 & 14 & 3 & 0 & 1 \\
    Zaragoza & 13 & 5 & 0 & 1 & 13 & 5 & 0 & 1 & 39 & 14 & 0 & 2 & 29 & 20 & 2 & 4 \\ \bottomrule
    \end{tabular}
    \caption{The original results of under-65 classifications}
    \label{tab:appunder65}
\end{table}